\documentclass{article} %
\usepackage{iclr2024_conference,times}

\usepackage{amsmath,amsfonts,bm}

\def\eqref#1{equation~\ref{#1}}

\def\1{\bm{1}}

\DeclareMathAlphabet{\mathsfit}{\encodingdefault}{\sfdefault}{m}{sl}
\SetMathAlphabet{\mathsfit}{bold}{\encodingdefault}{\sfdefault}{bx}{n}

\usepackage{graphicx, amsmath, amssymb, caption, subcaption, multirow, overpic, textpos, xspace}
\usepackage{enumitem}
\usepackage{booktabs}
\usepackage{xcolor}
\usepackage[british, english, american]{babel}
\usepackage[pagebackref=false, breaklinks=true, letterpaper=true, colorlinks,
            citecolor=citecolor, linkcolor=linkcolor, bookmarks=false]{hyperref}
\usepackage[capitalize]{cleveref}
\usepackage{gradient-text}

\definecolor{citecolor}{HTML}{0071BC}
\definecolor{linkcolor}{HTML}{ED1C24}

\newlength\savewidth

\renewcommand{\paragraph}[1]{\vspace{1.25mm}\noindent\textbf{#1}}

\newcommand{\app}{\raise.17ex\hbox{$\scriptstyle\sim$}}

\definecolor{deemph}{gray}{0.6}

\definecolor{baselinecolor}{gray}{.9}

\definecolor{brandeisblue}{rgb}{0.0, 0.44, 1.0}
\definecolor{darkpastelgreen}{rgb}{0.01, 0.75, 0.24}

\definecolor{brandeisblue}{rgb}{0.0, 0.44, 1.0}
\newcommand{\method}{{\texttt{EmerNeRF}}\xspace}

\setlist{noitemsep,leftmargin=*}

\title{\textsf{\gradientRGB{EmerNeRF}{254,50,254}{15,224,238}}: Emergent Spatial-Temporal Scene Decomposition via Self-Supervision}

\author{Jiawei Yang$^{*,\P}$, Boris Ivanovic$^{\P}$, Or Litany$^{\dag,\P}$, Xinshuo Weng$^{\P}$, Seung Wook Kim$^{\P}$, Boyi Li$^{\P}$, \\
\textbf{Tong Che}$^{\P}$, \textbf{Danfei Xu}$^{\$,\P}$, \textbf{Sanja Fidler}$^{\S,\P}$, \textbf{Marco Pavone}$^{\ddag,\P}$, \textbf{Yue Wang}$^{*,\P}$ \\
$^*$ \texttt{\{yangjiaw,yue.w\}@usc.edu}, University of Southern California \\ 
$^\$$ \texttt{danfei@gatech.edu}, Georgia Institute of Technology  \\
$^\S$ \texttt{fidler@cs.toronto.edu}, University of Toronto\\
$^\ddag$ \texttt{pavone@stanford.edu}, Stanford University\\
$^\dag$ \texttt{orlitany@gmail.com}, Technion\\
$^\P$ \texttt{\{bivanovic,xweng,seungwookk,boyil,tongc\}@nvidia.com}, NVIDIA Research
}

\iclrfinalcopy %
\begin{document}

\maketitle

\begin{abstract}

We present \method, a simple yet powerful approach for learning spatial-temporal representations of dynamic driving scenes. Grounded in neural fields, \method simultaneously captures scene geometry, appearance, motion, and semantics via self-bootstrapping. \method hinges upon two core components: First, it stratifies scenes into static and dynamic fields. This decomposition emerges purely from self-supervision, enabling our model to learn from general, in-the-wild data sources. Second, \method parameterizes an induced flow field from the dynamic field and uses this flow field to further aggregate multi-frame features, amplifying the rendering precision of dynamic objects. Coupling these three fields (static, dynamic, and flow) enables \method to represent highly-dynamic scenes self-sufficiently, without relying on ground truth object annotations or pre-trained models for dynamic object segmentation or optical flow estimation. Our method achieves state-of-the-art performance in sensor simulation, significantly outperforming previous methods when reconstructing static (+2.93 PSNR) and dynamic (+3.70 PSNR) scenes. In addition, to bolster \method's semantic generalization, we lift 2D visual foundation model features into 4D space-time and address a general positional bias in modern Transformers, significantly boosting 3D perception performance (e.g., 37.50\% relative improvement in occupancy prediction accuracy on average). Finally, we construct a diverse and challenging 120-sequence dataset to benchmark neural fields under extreme and highly-dynamic settings. 
See the project page for code, data, and request pre-trained models: {\small\url{https://emernerf.github.io}}
\end{abstract}
\section{Introduction}
\label{sec:intro}

\begin{figure}[ht!]
    \centering
    \includegraphics[width=1\linewidth]{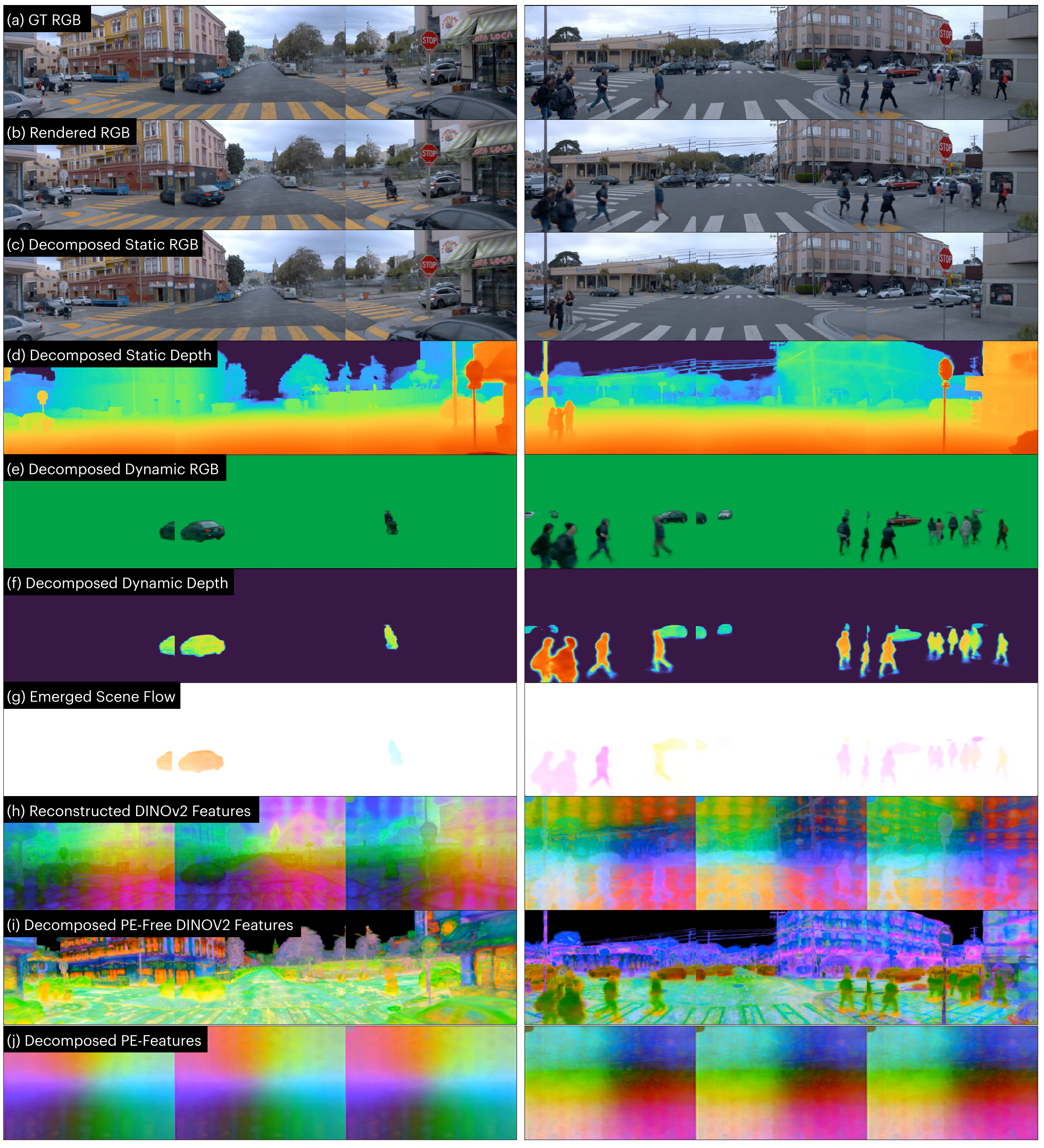}
    \caption{\method effectively reconstructs photo-realistic dynamic scenes (b), separating them into explicit static (c-d) and dynamic (e-f) elements, all via self-supervision. Notably, (g) scene flows emerge from \method without any explicit flow supervision. Moreover, \method can address detrimental positional embedding (PE) patterns observed in vision foundation models (h, j), and lift clean, PE-free features into 4D space (i). Additional visualizations can be found in \Cref{app:more_vis}.}
    \label{fig:teaser}
    \vspace{-1em}
\end{figure}

Perceiving, representing, and reconstructing dynamic scenes are critical for autonomous agents to understand and interact with their environments. Current approaches predominantly build custom pipelines with components dedicated to identifying and tracking static obstacles and dynamic objects~\citep{yang2023unisim,guo2023streetsurf}. However, such approaches require training each component with a large amount of labeled data and devising complex mechanisms to combine outputs across components.
To represent static scenes, approaches leveraging neural radiance fields (NeRFs)~\citep{mildenhall2021nerf} have witnessed a Cambrian explosion in computer graphics, robotics, and autonomous driving, owing to their strong performance in estimating 3D geometry and appearance~\citep{rematas2022urban,tancik2022block,wang2023neural,guo2023streetsurf}. However, without explicit supervision, NeRFs struggle with dynamic environments filled with
fast-moving objects, such as vehicles and pedestrians in urban scenarios. In this work, we tackle this long-standing challenge and develop a \emph{self-supervised} technique for building 4D (space-time) representations of dynamic scenes.

We consider a common setting where a mobile robot equipped with multiple sensors (e.g., cameras, LiDAR) navigates through a large dynamic environment (e.g., a neighborhood street). The fundamental underlying challenge is to construct an expansive 4D representation with only sparse, transient observations, that is, to reconstruct an entire 4D space-time volume from a \emph{single} traversal of the volume. Unfortunately, this setting challenges NeRF's multi-view consistency assumption---each point in the space-time volume will only be observed \emph{once}. Recent works~\citep{yang2023unisim,ost2021neural} seek to simplify the problem by modeling static scene components (e.g., buildings and trees) and dynamic objects separately, using a combination of neural fields and mesh representations. This decomposition enables exploiting multi-timestep observations to supervise static components, but it often requires costly ground-truth annotations to segment and track dynamic objects. Moreover, no prior works in this field have explicitly modeled temporal correspondence for dynamic objects, a prerequisite for accumulating sparse supervision over time. Overall, learning 4D representations of dynamic scenes remains a formidable challenge.

Towards this end, we present \method, a self-supervised approach for constructing 4D neural scene representations. As shown in~\cref{fig:teaser}, \method decouples static and dynamic scene components and estimates 3D scene flows --- remarkably, {\em all from self-supervision}.
At a high level, \method builds a hybrid static-dynamic world representation via a density-regularized objective, generating density for dynamic objects only as necessary (i.e., when points intersect dynamic objects). This representation enables our approach to capture dynamic components and exploit multi-timestep observations to self-supervise static scene elements. 
To address the lack of cross-observation consistency for dynamic components, we task \method to predict 3D scene flows and use them to aggregate temporally-displaced features. Intriguingly, \method's capability to estimate scene flow emerges naturally from this process, without any explicit flow supervision. Finally, to enhance scene comprehension, we ``lift'' features from pre-trained 2D visual foundation models (e.g., DINOv1~\citep{dinov1}, DINOv2~\citep{oquab2023dinov2}) to 4D space-time. In doing so, we observe and rectify a challenge tied to Transformer-based foundation models: positional embedding (PE) patterns (\cref{fig:teaser} (h)). As we will show in~\S\ref{sec:expt_feature_lifting}, effectively utilizing such general features greatly improves \method's semantic understanding and enables few-shot auto-labeling.

We evaluate \method on sensor sequences collected by autonomous vehicles (AVs) traversing through diverse urban environments. A critical challenge is that current autonomous driving datasets are heavily imbalanced, containing many simple scenarios with few dynamic objects. To facilitate a focused empirical study and bolster future research on this topic, we present the \textbf{N}eRF \textbf{O}n-\textbf{T}he-\textbf{R}oad (NOTR) benchmark, a balanced subsample of 120 driving sequences from the Waymo Open Dataset~\citep{sun2020scalability} containing diverse visual conditions (lighting, weather, and exposure) and challenging dynamic scenarios. On this benchmark, \method significantly outperforms previous state-of-the-art NeRF-based approaches~\citep{park2021hypernerf,wu2022d,muller2022instant,guo2023streetsurf} on scene reconstruction by 2.93 and 3.70 PSNR on static and dynamic scenes, respectively, and by 2.91 PSNR on dynamic novel view synthesis. For scene flow estimation, \method excels over \cite{li2021nsfp} by 42.16\% in metrics of interest. Additionally, removing PE patterns brings an average improvement of 37.50\% relative to using the original, PE pattern-laden features on semantic occupancy prediction.
\textbf{Contributions.} Our key contributions are fourfold:
(1) We introduce \method, a novel 4D neural scene representation framework that excels in challenging autonomous driving scenarios. \method performs static-dynamic decomposition and scene flow estimation, all through self-supervision. 
(2) A streamlined method to tackle the undesired effects of positional embedding patterns from Vision Transformers, which is immediately applicable to other tasks. 
(3) We introduce the NOTR dataset to assess neural fields in diverse conditions and facilitate future development in the field. 
(4) \method achieves state-of-the-art performance in scene reconstruction, novel view synthesis, and scene flow estimation. 
\section{Related Work}

\textbf{Dynamic scene reconstruction with NeRFs.} Recent works adopt NeRFs~\citep{mildenhall2021nerf,muller2022instant} to accommodate dynamic scenes~\citep{li2021nsff,park2021hypernerf,wu2022d}. Earlier methods \citep{bansal20204d,li2022neural,wang2022fourier,fang2022fast} for dynamic view synthesis rely on multiple synchronized videos recorded from different viewpoints, restricting their use for real-world applications in autonomous driving and robotics. Recent methods, such as Nerfies \citep{park2021nerfies} and HyperNeRF \citep{park2021hypernerf}, have managed to achieve dynamic view synthesis using a single camera. However, they rely on a strict assumption that all observations can be mapped via deformation back to a canonical reference space, usually constructed from the first timestep. This assumption does not hold in driving because objects might not be fully present in any single frame and can constantly enter and exit the scene. %

Of particular relevance to our work are methods like D$^2$NeRF \citep{wu2022d}, SUDS \citep{turki2023suds}, and NeuralGroundplans \citep{sharma2022neural}. These methods also partition a 4D scene into static and dynamic components. However, D$^2$NeRF underperforms significantly for outdoor scenes due to its sensitivity to hyperparameters and insufficient capacity; NeuralGroundplan relies on synchronized videos from different viewpoints to reason about dynamics; and SUDS, designed for multi-traversal driving logs, largely relies on accurate optical flows derived by pre-trained models and incurs high computational costs due to its expensive flow-based warping losses. In contrast, our approach can reconstruct an accurate 4D scene representation from a single-traversal log captured by sensors mounted on a self-driving vehicle. Freed from the constraints of pre-trained flow models, \method exploits and refines its own intrinsic flow predictions, enabling a self-improving loop.

\textbf{NeRFs for AV data.} Creating high-fidelity neural simulations from collected driving logs is crucial for the autonomous driving community, as it facilitates the closed-loop training and testing of various algorithms. Beyond SUDS~\citep{turki2023suds}, there is a growing interest in reconstructing scenes from driving logs. 
In particular, recent methods excel with static scenes but face challenges with dynamic objects~\citep{guo2023streetsurf}. While approaches like UniSim~\citep{yang2023unisim} and NSG~\citep{ost2021neural} handle dynamic objects, they depend on ground truth annotations, making them less scalable due to the cost of obtaining such annotations. In contrast, our method achieves high-fidelity simulation results purely through self-supervision, offering a scalable solution.

\textbf{Augmenting NeRFs.} NeRF methods are commonly augmented with external model outputs to incorporate additional information. For example, approaches that incorporate scene flow often rely on existing optical flow models for supervision \citep{li2021nsff,turki2023suds,li2023dynibar}. They usually require cycle-consistency tests to filter out inconsistent flow estimations; otherwise, the optimization process is prone to failure \citep{wang2023tracking}. The Neural Scene Flow Prior (NSFP) \citep{li2021nsfp}, a state-of-the-art flow estimator, optimizes a neural network to estimate the scene flow at each timestep (minimizing the Chamfer Loss \citep{fan2017point}). This per-timestep optimization makes NSFP prohibitively expensive. In contrast, our \method bypasses the need for either pre-trained optical flow models or holistic geometry losses. Instead, our flow field is supervised only by scene reconstruction losses and the flow estimation capability \textit{emerges on its own}.
Most recently, 2D signals such as semantic labels or foundation model feature vectors have been distilled into 3D space~\citep{kobayashi2022decomposing,lerf2023,tsagkas2023vlfields,shafiullah2022clipfields}, enabling semantic understanding tasks. In this work, we similarly lift visual foundation model features into 4D space and show their potential for few-shot perception tasks.

\section{Self-Supervised Spatial-Temporal Neural Fields}
\label{sec:method}
Learning a spatial-temporal representation of a dynamic environment with a multi-sensor robot is challenging due to the sparsity of observations and costs of obtaining ground truth annotations.
To this end, our design choices stem from the following key principles: (1) Learn a scene decomposition entirely through self-supervision and avoid using any ground-truth annotations or pre-trained models for dynamic object segmentation or optical flow. (2) Model dynamic element correspondences across time via scene flow. (3) Obtain a mutually reinforcing representation: static-dynamic decomposition and flow estimation can benefit from each other. (4) Improve the semantics of scene representations by leveraging feature lifting and distillation, enabling a range of perception tasks. 

Having established several design principles, we are now equipped to describe \method, a self-supervised approach for efficiently representing both static and dynamic scene components. First, \S\ref{sec:scene_encoder} details how \method builds a hybrid world representation with a static and dynamic field. Then, \S\ref{sec:flow_encoder} explains how \method leverages an emergent flow field to aggregate temporal features over time, further improving its representation of dynamic components. \S\ref{sec:vit_feature_lifting} describes the lifting of semantic features from pre-trained 2D models to 4D space-time, enhancing \method's scene understanding. Finally, \S\ref{sec:optimization} discusses the loss function that is minimized during training. 

\subsection{Scene Representations}
\label{sec:scene_encoder}
\begin{figure}
    \centering
    \includegraphics[width=1\linewidth]{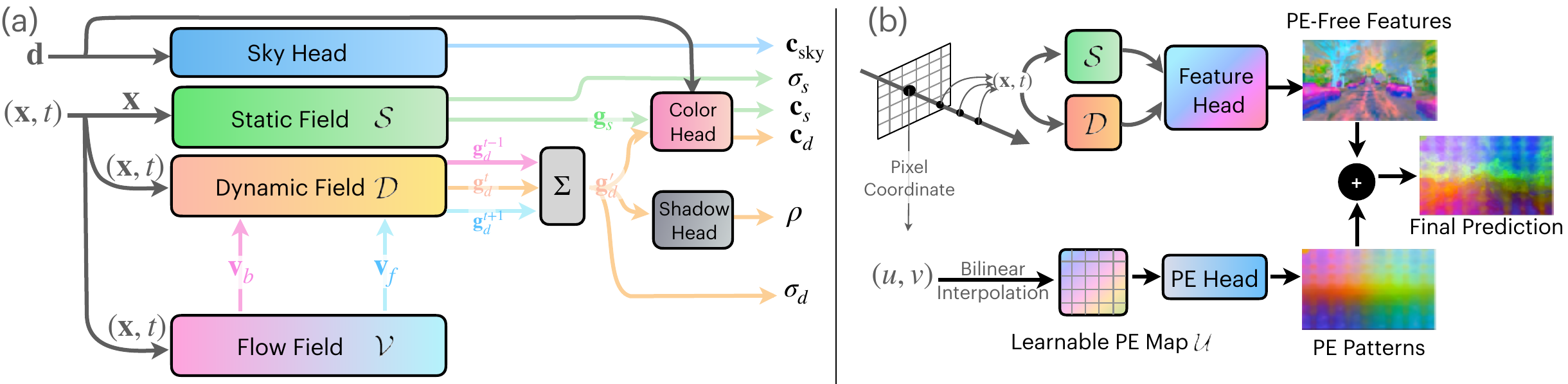}
    \caption{\textbf{\method Overview.} (a) \method consists of a static, dynamic, and flow field ($\mathcal{S}, \mathcal{D}, \mathcal{V}$). These fields take as input either a spatial query $\mathbf{x}$ or spatial-temporal query $(\mathbf{x}, t)$ to generate a static (feature $\mathbf{g}_s$, density $\sigma_s$) pair or a dynamic (feature $\mathbf{g}_d'$, density $\sigma_d$) pair. Of note, we use the forward and backward flows ($\mathbf{v}_f$ and $\mathbf{v}_b$) to generate temporally-aggregated features $\mathbf{g}_d'$ from nearby temporal features $\mathbf{g}_{d}^{t-1}$, $\mathbf{g}_{d}^{t}$, and $\mathbf{g}_{d}^{t+1}$ (a slight abuse of notation w.r.t. \cref{eq:temp_feats}). These features (along with the view direction $\mathbf{d}$) are consumed by the shared color head which independently predicts the static and dynamic colors $\mathbf{c}_s$ and $\mathbf{c}_d$.
    The shadow head predicts a shadow ratio $\rho$ from the dynamic features. The sky head decodes a per-ray color $\mathbf{c}_\mathrm{sky}$ for sky pixels from the view direction $\mathbf{d}$. (b) \method renders the aggregated features to 2D and removes undesired positional encoding patterns (via a learnable PE map followed by a lightweight PE head).}
 
    \label{fig:overview}
\end{figure}
\textbf{Scene decomposition.} To enable efficient scene decomposition, we design \method to be a hybrid spatial-temporal representation. It decomposes a 4D scene into a static field $\mathcal{S}$ and a dynamic field $\mathcal{D}$, both of which are parameterized by learnable hash grids~\citep{muller2022instant} $\mathcal{H}_s$ and $\mathcal{H}_d$, respectively. This decoupling offers a flexible and compact 4D scene representation for time-independent features $\mathbf{h}_s=\mathcal{H}_s(\mathbf{x})$ and time-varying features $\mathbf{h}_d=\mathcal{H}_d(\mathbf{x}, t)$, where $\mathbf{x}=(x, y, z)$ is the 3D location of a query point and $t$ denotes its timestep. These features are further transformed into $ \mathbf{g}_s$ and $\mathbf{g}_d$  by  lightweight MLPs ($g_s$ and $g_d$) and used to predict per-point density $\sigma_s$ and $\sigma_d$: 
\begin{align}
    \mathbf{g}_s, \sigma_s &= g_s(\mathcal{H}_s(\mathbf{x})) & \mathbf{g}_d, \sigma_d &= g_d(\mathcal{H}_d(\mathbf{x}, t))
\end{align}

\textbf{Multi-head prediction.} \method uses separate heads for color, sky, and shadow predictions. 
To maximize the use of dense self-supervision from the static branch, the static and dynamic branches share the same color head $\mathrm{MLP}_{\mathrm{color}}$. This color head takes ($\mathbf{g}_s$, $\mathbf{d}$) and ($\mathbf{g}_d$, $\mathbf{d}$) as input, and outputs per-point color $\mathbf{c}_s$ and $\mathbf{c}_d$ for static and dynamic items, where $\mathbf{d}$ is the normalized view direction. Since the depth of the sky is ill-defined, we follow \cite{rematas2022urban} and use a separate sky head to predict the sky's color from the frequency-embedded view direction $\gamma(\mathbf{d})$
, where $\gamma(\cdot)$ is a frequency-based positional embedding function, as in \cite{mildenhall2021nerf}. 
Lastly, as in \cite{wu2022d}, we use a shadow head $\mathrm{MLP}_{\mathrm{shadow}}$ to depict the shadows of dynamic objects. It outputs a scalar $\rho \in [0, 1]$ for dynamic objects, modulating the color intensity predicted by the static field. Collectively, we have:
\begin{align}
    \mathbf{c}_s &= \mathrm{MLP}_{\mathrm{color}}(\mathbf{g}_s, \gamma(\mathbf{d})) & \mathbf{c}_d &= \mathrm{MLP}_{\mathrm{color}}(\mathbf{g}_d, \gamma(\mathbf{d})) \label{eq:color} \\
    \mathbf{c}_{\mathrm{sky}} &= \mathrm{MLP}_{\mathrm{color\_sky}}(\gamma(\mathbf{d})) & \rho &= \mathrm{MLP}_{\mathrm{shadow}}(\mathbf{g}_d) \label{eq:sky} 
\end{align}

\textbf{Rendering.} To enable highly-efficient rendering, we use density-based weights to combine results from the static field and dynamic field:
\begin{align}
    \mathbf{c} &= \frac{\sigma_s}{\sigma_s + \sigma_d} \cdot (1 - \rho)\cdot \mathbf{c}_s + \frac{\sigma_d}{\sigma_s + \sigma_d} \cdot\mathbf{c}_d
    \label{eq:color_rendering}
\end{align}
To render a pixel, we use $K$ discrete samples $\{\mathbf{x}_1, \dots, \mathbf{x}_K\}$ along its ray to estimate the integral of color. The final outputs are given by:
\begin{align}
    \hat{C} = \sum_{i=1}^{K} T_i \alpha_i \mathbf{c}_i + \left(1 - \sum_{i=1}^{K} T_i \alpha_i\right)\mathbf{c}_\mathrm{sky} \label{eq:render}
\end{align}
where $T_i = \prod_{j=1}^{i-1}(1-\alpha_j)$ is the accumulated transmittance and $\alpha_i= 1- \exp(-\sigma_i (\mathbf{x}_{i+1} - \mathbf{x}_i))$ is the piece-wise opacity.

\textbf{Dynamic density regularization.} 
To facilitate static-dynamic decomposition, we leverage the fact that our world is predominantly static. We regularize dynamic density by minimizing the expectation of the dynamic density $\sigma_d$, which prompts the dynamic field to produce density values only as needed: 
\begin{equation}
    \mathcal{L}_{\sigma_d} = \mathbb{E}(\sigma_d)
\label{eq:density_reg}
\end{equation}

\subsection{Emergent Scene Flow}
\label{sec:flow_encoder}
\textbf{Scene flow estimation.} To capture explicit correspondences between dynamic objects and provide a link by which to aggregate temporally-displaced features, we introduce an additional scene flow field consisting of a hash grid $\mathcal{V} := \mathcal{H}_v(\mathbf{x}, t)$ and a flow predictor $\mathrm{MLP}_v$. 
This flow field maps a spatial-temporal query point $(\mathbf{x},t)$ to a flow vector $\mathbf{v}\in \mathbb{R}^3$, which transforms the query point to its position in the next timestep, given by: 
\begin{align}
   \mathbf{v} &= \mathrm{MLP}_v(\mathcal{H}_v(\mathbf{x}, t)) & \mathbf{x}'&= \mathbf{x} + \mathbf{v}
\end{align}
In practice, our flow field predicts both a forward flow $\mathbf{v}_f$ and a backward flow $\mathbf{v}_b$, resulting in a 6-dimensional flow vector for each point. 

\textbf{Multi-frame feature integration.} Next, we use the link provided by the predicted scene flow to integrate features from nearby timesteps, using a simple weighted summation:
\begin{equation}
    \mathbf{g}_d' = 0.25 \cdot g_d(\mathcal{H}_d(\mathbf{x} + \mathbf{v}_b, t-1)) + 0.5 \cdot g_d(\mathcal{H}_d(\mathbf{x}, t)) + 0.25 \cdot g_d(\mathcal{H}_d(\mathbf{x} + \mathbf{v}_f, t+1))
    \label{eq:temp_feats}
\end{equation}
If not otherwise specified, $\mathbf{g}_d'$ is used by default when the flow field is enabled (instead of $\mathbf{g}_d$ in \cref{eq:color,eq:sky}).
This feature aggregation module achieves three goals: 1) It connects the flow field to scene reconstruction losses (e.g., RGB loss) for supervision, 2) it consolidates features, denoising temporal attributes for accurate predictions, and 3) each point is enriched through the shared gradient of its temporally-linked features, enhancing the quality of individual points via shared knowledge.

\textbf{Emergent abilities.} We do not use any explicit flow supervision to guide \method's flow estimation process. Instead, this capability \textit{emerges} from our temporal aggregation step while optimizing scene reconstruction losses (\S\ref{sec:optimization}). Our hypothesis is that only temporally-consistent features benefit from multi-frame feature integration, and this integration indirectly drives the scene flow field toward optimal solutions --- predicting correct flows for all points. Our subsequent ablation studies in \cref{app:ablation} confirm this: when the temporal aggregation is disabled or gradients of these nearby features are stopped, the flow field fails to learn meaningful results.

\subsection{Vision Transformer Feature Lifting}
\label{sec:vit_feature_lifting}
While NeRFs excel at generating high-fidelity color and density fields, they lack in conveying semantic content, constraining their utility for semantic scene comprehension. To bridge this gap, we lift 2D foundation model features to 4D, enabling crucial autonomous driving perception tasks such as semantic occupancy prediction. Although previous works might suggest a straightforward approach \citep{lerf2023,kobayashi2022decomposing}, directly lifting features from state-of-the-art vision transformer (ViT) models has hidden complexities due to positional embeddings (PEs) in transformer models (\cref{fig:teaser} (h-j)). In the following sections, we detail how we enhance \method with a feature reconstruction head, uncover detrimental PE patterns in transformer models, and subsequently mitigate these issues.

\textbf{Feature reconstruction head.}  Analogous to the color head, we incorporate a feature head $\mathrm{MLP}_\mathrm{feat}$ and a feature sky head $\mathrm{MLP}_\mathrm{feat\_sky}$ to predict per-point features  $\mathbf{f}$ and sky features $\mathbf{f}_\mathrm{sky}$, given by:
\begin{align}
    \mathbf{f}_{*} &= \mathrm{MLP}_\mathrm{feat}(\mathbf{g_{*}}), \text{ where } * \in \{s, d\} & \mathbf{f}_\mathrm{sky} = \mathrm{MLP}_\mathrm{feat\_sky}(\gamma(\mathbf{d})). 
\end{align}
Similar to the color head, we share the feature head among the static and dynamic branches.
Rendering these features similarly follows~\cref{eq:render}, given by:
\begin{equation}
    \hat{F} = \sum_{i=1}^{K} T_i \alpha_i \mathbf{f}_i + \left(1 - \sum_{i=1}^{K} T_i \alpha_i\right)\mathbf{f}_\mathrm{sky} \label{eq:feats}    
\end{equation}

\textbf{Positional embedding patterns.} We observe pronounced and undesired PE patterns when using current state-of-the-art foundation models, notably DINOv2~\citep{oquab2023dinov2} (\cref{fig:teaser} (h)). These patterns remain fixed in images, irrespective of 3D viewpoint changes, breaking 3D multi-view consistency. 
Our experiments (\S\ref{sec:expt_feature_lifting}) reveal that these patterns not only impair feature synthesis results, but also cause a substantial reduction in 3D perception performance.

\textbf{Shared learnable additive prior.}
We base our solution on the observation that ViTs extract feature maps image-by-image and these PE patterns appear (almost) consistently across all images. This suggests that a single PE feature map might be sufficient to capture this shared artifact. Accordingly, we assume an additive noise model for the PE patterns; that is, they can be independently subtracted from the original features to obtain PE-free features. With this assumption, we construct a learnable and globally-shared 2D feature map $\mathcal{U}$ to compensate for these patterns. This process is depicted in~\cref{fig:overview} (b). For a target pixel coordinate $(u, v)$, we first volume-render a PE-free feature as in~\cref{eq:feats}. Then, we bilinearly interpolate $\mathcal{U}$ and decode the interpolated feature using a single-layer $\mathrm{MLP}_{\mathrm{PE}}$ to obtain the PE pattern feature, which is then added to the PE-free feature. Formally:
\begin{equation}
    \hat{F} = \underbrace{\sum_{i=1}^{K} T_i \alpha_i \mathbf{f}_i + \left(1 - \sum_{i=1}^{k} T_i \alpha_i\right)\mathbf{f}_\mathrm{sky}}_{\text{Volume-rendered PE-free feature}} + \underbrace{\mathrm{MLP}_{\mathrm{PE}}\left(\texttt{interp}\left(\left(u, v\right), \mathcal{U}\right)\right)}_{\text{PE feature}}
\end{equation}
The grouped terms render ``PE-free'' features (\cref{fig:teaser} (i)) and ``PE'' patterns (\cref{fig:teaser} (j)), respectively, with their sum producing the overall ``PE-containing'' features (\cref{fig:teaser} (h)).

\subsection{Optimization}
\label{sec:optimization}

\textbf{Loss functions.} 
Our method decouples pixel rays and LiDAR rays to account for sensor asynchronization. For pixel rays, we use an L2 loss for colors $\mathcal{L}_\mathrm{rgb}$ (and optional semantic features $\mathcal{L}_\mathrm{feat}$), a binary cross entropy loss for sky supervision $\mathcal{L}_\mathrm{sky}$, and a shadow sparsity loss $\mathcal{L}_\mathrm{shadow}$. For LiDAR rays, we combine an expected depth loss with a line-of-sight loss $\mathcal{L}_\mathrm{depth}$, as proposed in \cite{rematas2022urban}. This line-of-sight loss promotes an unimodal distribution of density weights along a ray, which we find is important for clear static-dynamic decomposition. For dynamic regularization, we use a density-based regularization (Eq.~\ref{eq:density_reg}) to encourage the dynamic field to produce density values only when absolutely necessary. This dynamic regularization loss is applied to both pixel rays ($\mathcal{L}_{\sigma_d\text{(pixel)}}$) and LiDAR rays ($\mathcal{L}_{\sigma_d(\text{LiDAR})}$). Lastly, we regularize the flow field with a cycle consistency loss $\mathcal{L}_{\text{cycle}}$. See \cref{app:specific_impl_details} for details. In summary, we minimize:
\begin{equation}
    \mathcal{L} = \underbrace{\mathcal{L}_{\text{rgb}} + \mathcal{L}_{\text{sky}} + \mathcal{L}_{\text{shadow}} + \mathcal{L}_{\sigma_d\text{(pixel)}} + \mathcal{L}_{\text{cycle}} + \mathcal{L}_{\text{feat}}}_{\text{for pixel rays}} + \underbrace{\mathcal{L}_{\text{depth}} + \mathcal{L}_{\sigma_d(\text{LiDAR})}}_{\text{for LiDAR rays}} %
    \label{eq:total_loss_fn}
\end{equation}

\textbf{Implementation details.} All model implementation details can be found in~\cref{app:implementation}. %

\section{Experiments}
\label{sec:experiment}
In this section, we benchmark the reconstruction capabilities of \method against prior methods, focusing on static and dynamic scene reconstruction, novel view synthesis, scene flow estimation, and foundation model feature reconstruction. Further ablation studies and a discussion of \method's limitations can be found in \cref{app:ablation,app:limitations}, respectively.

\textbf{Dataset.} While there exist many public datasets with AV sensor data~\citep{CaesarBankitiEtAl2019,sun2020scalability,nuplan2021}, they are heavily imbalanced, containing many simple scenarios with few to no dynamic objects. To remedy this, we introduce \textbf{N}eRF \textbf{O}n-\textbf{T}he-\textbf{R}oad (NOTR), a balanced and diverse benchmark derived from the Waymo Open Dataset \citep{sun2020scalability}. NOTR features 120 unique, hand-picked driving sequences, split into 32 static (the same split as in StreetSurf~\citep{guo2023streetsurf}), 32 dynamic, and 56 diverse scenes across seven challenging conditions: ego-static, high-speed, exposure mismatch, dusk/dawn, gloomy, rainy, and night. We name these splits Static-32, Dynamic-32, and Diverse-56, respectively. This dataset not only offers a consistent benchmark for static and dynamic object reconstruction, it also highlights the challenges of training NeRFs on real-world AV data. Beyond simulation, our benchmark offers 2D bounding boxes for dynamic objects, ground truth 3D scene flow, and 3D semantic occupancy---all crucial for driving perception tasks.
Additional details can be found in~\cref{app:dataset}.
\subsection{Rendering} 

\textbf{Setup.} To analyze performance across various driving scenarios, we test \method's scene reconstruction and novel view synthesis capabilities on different NOTR splits. For scene reconstruction, all samples in a log are used for training. This setup probes the upper bound of each method. For novel view synthesis, we omit every 10th timestep, resulting in 10\% novel temporal views for evaluation. Our metrics include peak signal-to-noise ratio (PSNR) and structural similarity index (SSIM). For dynamic scenes, we further leverage ground truth bounding boxes and velocity data to identify dynamic objects and compute ``dynamic-only'' metrics; and we benchmark against HyperNeRF \citep{park2021hypernerf} and D$^2$NeRF \citep{wu2022d}, two state-of-the-art methods for modeling dynamic scenes. Due to their prohibitive training cost, we only compare against them in the Dynamic-32 split.
On the Static-32 split, we disable our dynamic and flow branches, and compare against StreetSurf \citep{guo2023streetsurf} and iNGP \citep{muller2022instant} (as implemented by \cite{guo2023streetsurf}). We use the official codebases released by these methods, and adapt them to NOTR. To ensure a fair comparison, we augment all methods with LiDAR depth supervision and sky supervision, and disable our feature field. Further details can be found in \cref{app:baselines}.

\begin{table}[tb]
\centering
\caption{\textbf{Dynamic and static scene reconstruction performance.}} 
\label{tab:combined}
\begin{minipage}{0.65\textwidth}
\centering
(a) Dynamic-32 Split \\
\resizebox{\linewidth}{!}{%
\begin{tabular}{lccccccccc}
\toprule
\multirow{3}{*}{Methods} & \multicolumn{4}{c}{Scene Reconstruction}      & \multicolumn{4}{c}{Novel View Synthesis}   \\ 
\cmidrule(lr){2-5} \cmidrule(lr){6-9} 
 & \multicolumn{2}{c}{Full Image}           & \multicolumn{2}{c}{Dynamic-Only} & \multicolumn{2}{c}{Full Image}           & \multicolumn{2}{c}{Dynamic-Only} \\
 & PSNR$\uparrow$ & SSIM$\uparrow$  & PSNR$\uparrow$      & SSIM$\uparrow$      & PSNR$\uparrow$ & SSIM$\uparrow$ & DPSNR$\uparrow$     & SSIM$\uparrow$     \\ 
\midrule
D$^2$NeRF   & 24.35 & 0.645  & 21.78  & 0.504  & 24.17  & 0.642 & 21.44  & 0.494 \\
HyperNeRF   & 25.17 & 0.688  & 22.93  & 0.569  & 24.71  & 0.682 &  22.43 & 0.554\\
\midrule 
Ours & \textbf{28.87} & \textbf{0.814}& \textbf{26.19} & \textbf{0.736} &  \textbf{27.62} & \textbf{0.792} & \textbf{24.18} & \textbf{0.670}
\\

\bottomrule
\end{tabular}%
}
\end{minipage}
\hfill
\begin{minipage}{0.33\textwidth}
\centering
(b) Static-32 Split \\
\resizebox{\linewidth}{!}{%
\begin{tabular}{lcc}
\toprule
\multirow{2}{*}{Methods} & \multicolumn{2}{c}{Static Scene Reconstruction}     \\ 
\cmidrule(lr){2-3} 
 & PSNR$\uparrow$ & SSIM$\uparrow$  \\ 
\midrule
iNGP    & 24.46    & 0.694  \\
StreetSurf & 26.15 & 0.753      \\
\midrule 
Ours & \textbf{29.08} & \textbf{0.803} \\
\bottomrule
\end{tabular}%
}
\end{minipage}

\vspace{-0.2cm}

\end{table}

\textbf{Dynamic scene comparisons.} \cref{tab:combined} (a) shows that our approach consistently outperforms others on scene reconstruction and novel view synthesis. We refer readers to \cref{app:qualitative_comparison} for qualitative comparisons. In them, we can see that HyperNeRF \citep{park2021hypernerf} and D$^2$NeRF \citep{wu2022d} tend to produce over-smoothed renderings and struggle with dynamic object representation. In contrast, \method excels in reconstructing high-fidelity static background and dynamic foreground objects, while preserving high-frequency details (evident from its high SSIM and PSNR values). Despite D$^2$NeRF's intent to separate static and dynamic elements, it struggles in complex driving contexts and produces poor dynamic object segmentation (as shown in \cref{fig:dynamic_nvs}). Our method outperforms them both quantitatively and qualitatively. 
\textbf{Static scene comparisons.} While static scene representation is not our main focus, \method excels in this aspect too, as evidenced in \cref{tab:combined} (b). It outperforms state-of-the-art StreetSuRF \citep{guo2023streetsurf} which is designed for static outdoor scenes. With the capability to model both static and dynamic components, \method can accurately represent more general driving scenes.

\subsection{Flow Estimation} 

\textbf{Setup.} We assess \method on all frames of the Dynamic-32 split, benchmarking against the prior state-of-the-art, NSFP \citep{li2021nsfp}. Using the Waymo dataset's ground truth scene flows, we compute metrics consistent with \cite{li2021nsfp}: 3D end-point error (EPE3D), calculated as the mean L2 distance between predictions and ground truth for all points; $\mathrm{Acc}_5$, representing the fraction of points with EPE3D less than 5cm or a relative error under 5\%; $\mathrm{Acc}_{10}$, indicating the fraction of points with EPE3D under 10cm or a relative error below 10\%; and $\theta$, the average angle error between predictions and ground truths. When evaluating NSFP \citep{li2021nsfp}, we use their official implementation and remove ground points (our approach does not require such preprocessing).

\textbf{Results.} As shown in \cref{tab:scene_flow_estimation}, our approach outperforms NSFP across all metrics, with significant leads in EPE3D, $\mathrm{Acc}_5$, and $\mathrm{Acc}_{10}$. While NSFP \citep{li2021nsfp} employs the Chamfer distance loss \cite{fan2017point} to solve scene flow, \method achieves significantly better results without any explicit flow supervision. These properties naturally emerge from our temporal aggregation step. \cref{app:ablation} contains additional ablation studies regarding the emergence of flow estimation.

\begin{table}[tb]
\centering
\caption{\textbf{Scene flow estimation on the NOTR Dynamic-32 split.}}
\label{tab:scene_flow_estimation}
\resizebox{0.7\textwidth}{!}{%
\begin{tabular}{lcccc}
\toprule
Methods & EPE3D $ (m) \downarrow$ & $\text{Acc}_{5} (\%)$  $\uparrow$ & $\text{Acc}_{10} (\%) \uparrow$ &  $\theta$ (rad) $\downarrow$ \\ \midrule
NSFP \citep{li2021nsfp}      & 0.365    & 51.76  & 67.36 & 0.84  \\
Ours & \textbf{0.014} & \textbf{93.92}  & \textbf{96.27}  & \textbf{0.64}  \\
\bottomrule
\end{tabular}%
}

\vspace{-0.25cm}

\end{table}

\begin{table}[tb]
\centering
\caption{\textbf{Few-shot semantic occupancy prediction evaluation.} We investigate the influence of positional embedding (PE) patterns on 4D features by evaluating semantic occupancy prediction performance. We report sample-averaged micro-accuracy and class-averaged macro-accuracy.}
\label{tab:occ_eval}
\resizebox{\textwidth}{!}{%
\begin{tabular}{llcccccccc}
\toprule
\multirow{2}{*}{PE removed?} & \multirow{2}{*}{ViT model} & \multicolumn{2}{c}{Static-32} & \multicolumn{2}{c}{Dynamic-32} & \multicolumn{2}{c}{Diverse-56}  & \multicolumn{2}{c}{Average of 3 splits} \\ 
\cmidrule(lr){3-4}  \cmidrule(lr){5-6} \cmidrule(lr){7-8} \cmidrule(lr){9-10}
 &  & Micro Acc & Macro Acc & Micro Acc & Macro Acc & Micro Acc & Macro Acc & Micro Acc & Macro Acc \\ 
\midrule 
No & DINOv1 & 43.12\% & 52.71\% & 47.51\% & 54.46\% & 43.19\% & 51.11\% & 44.60\% & 52.76\% \\
Yes & DINOv1 & \textbf{55.02\%} & \textbf{57.13\%} & \textbf{57.65\%} & \textbf{57.77\%} & \textbf{54.56\%} & \textbf{55.13\%} & \textbf{55.74\%} & \textbf{56.67\%} \\
\multicolumn{2}{l}{Relative Improvement} &  {\footnotesize \color{darkpastelgreen}+27.60\%} & {\footnotesize \color{darkpastelgreen}+8.38\%} & {\footnotesize \color{darkpastelgreen}+21.35\%} & {\footnotesize \color{darkpastelgreen}+6.07\%} & {\footnotesize \color{darkpastelgreen}+26.32\%} & {\footnotesize \color{darkpastelgreen}+7.87\%} & {\footnotesize \color{darkpastelgreen}+24.95\%} & {\footnotesize \color{darkpastelgreen}+7.42\%} \\ \midrule

No & DINOv2 & 38.73\% & 50.30\% & 51.43\% & 57.03\% & 45.22\% & 54.37\% & 45.13\% & 53.90\% \\
Yes & DINOv2 & \textbf{63.21\%} & \textbf{59.41\%} & \textbf{65.08\%} & \textbf{60.82\%} & \textbf{57.86\%} & \textbf{59.00\%} & \textbf{62.05\%} & \textbf{59.74\%} \\
\multicolumn{2}{l}{Relative Improvement} &  {\footnotesize \color{darkpastelgreen}+63.22\%} & {\footnotesize \color{darkpastelgreen}+18.11\%} & {\footnotesize \color{darkpastelgreen}+26.53\%} & {\footnotesize \color{darkpastelgreen}+6.65\%} & {\footnotesize \color{darkpastelgreen}+27.95\%} & {\footnotesize \color{darkpastelgreen}+8.51\%} & {\footnotesize \color{darkpastelgreen}+37.50\%} & {\footnotesize \color{darkpastelgreen}+10.84\%} \\
\bottomrule
\end{tabular}%
}
\end{table}

\subsection{Leveraging Foundation Model Features}\label{sec:expt_feature_lifting}

To investigate the impact of ViT PE patterns on 3D perception and feature synthesis, we instantiate versions of \method with and without our proposed PE decomposition module. 

\textbf{Setup.} We evaluate \method's few-shot perception capabilities using the Occ3D dataset \citep{tian2023occ3d}. Occ3D provides 3D semantic occupancy annotations for the Waymo dataset \citep{sun2020scalability} in voxel sizes of 0.4m and 0.1m (we use 0.1m). For each sequence, we annotate every 10th frame with ground truth information, resulting in 10\% labeled data. Occupied coordinates are input to pre-trained \method models to compute feature centroids per class. Features from the remaining 90\% of frames are then queried and classified based on their nearest feature centroid. We report both micro (sample-averaged) and macro (class-averaged) classification accuracies. All models are obtained from the scene reconstruction setting, i.e., all views are used for training.

\textbf{Results.} Table \ref{tab:occ_eval} compares the performance of PE-containing 4D features to their PE-free counterparts. Remarkably, \method with PE-free DINOv2~\citep{oquab2023dinov2} features sees a maximum relative improvement of 63.22\% in micro-accuracy and an average increase of 37.50\% over its PE-containing counterpart. Intriguingly, although the DINOv1~\citep{dinov1} model might appear visually unaffected (\cref{fig:pos_patterns}), our results indicate that directly lifting PE-containing features to 4D space-time is indeed problematic. With our decomposition, PE-free DINOv1 features witness an average relative boost of 24.95\% in micro-accuracy. 
As another illustration of PE patterns' impact, by eliminating PE patterns, the improved performance of DINOv2 over DINOv1 carries over to 3D perception (e.g., Static32 micro-accuracy).
\begin{table}[tb]
\centering
\caption{\textbf{Feature synthesis results.} We report the feature-PNSR values under different settings.}
\label{tab:feat_synthesis}
\resizebox{0.6\textwidth}{!}{%
\begin{tabular}{rllll}
\toprule
PE removed? & ViT model & Static-32 & Dynamic-32 & Diverse-56  \\ 
\midrule 
No & DINOv1 & 23.35 & 23.37 & 23.78 \\
Yes & DINOv1 & 23.57 {\footnotesize \color{darkpastelgreen}(+0.23)} & 23.52 {\footnotesize \color{darkpastelgreen}(+0.15)} & 23.92 {\footnotesize \color{darkpastelgreen}(+0.14)} \\
\midrule 
No & DINOv2 & 21.87 & 22.34 & 22.79 \\
Yes & DINOv2 & 22.70 {\footnotesize \color{darkpastelgreen}(+0.83)} & 22.80 {\footnotesize \color{darkpastelgreen}(+0.45)} & 23.21 {\footnotesize \color{darkpastelgreen}(+0.42)} \\
\bottomrule
\end{tabular}%
}
\end{table}

\textbf{Feature synthesis results.} Table \ref{tab:feat_synthesis} compares the feature-PSNR of PE-containing and PE-free models, showing marked improvements in feature synthesis quality when using our proposed PE decomposition method, especially for DINOv2~\citep{oquab2023dinov2}. While DINOv1~\citep{dinov1} appears to be less influenced by PE patterns, our method unveils their presence, further showing that even seemingly unaffected models can benefit from PE pattern decomposition.

\section{Conclusion}

In this work, we present \method, a simple yet powerful approach for learning 4D neural representations of dynamic scenes. \method effectively captures scene geometry, appearance, motion, and any additional semantic features by decomposing scenes into static and dynamic fields, learning an emerged flow field, and optionally lifting foundation model features to a resulting hybrid world representation. \method additionally removes problematic positional embedding patterns that appear when employing Transformer-based foundation model features. Notably, all of these tasks (save for foundation model feature lifting) are learned in a \emph{self-supervised} fashion, without relying on ground truth object annotations or pre-trained models for dynamic object segmentation or optical flow estimation. When evaluated on NOTR, our carefully-selected subset of 120 challenging driving scenes from the Waymo Open Dataset~\citep{sun2020scalability}, \method achieves state-of-the-art performance in sensor simulation, significantly outperforming previous methods on both static and dynamic scene reconstruction, novel view synthesis, and scene flow estimation. Exciting areas of future work include further exploring capabilities enabled or significantly improved by harnessing foundation model features: few-shot, zero-shot, and auto-labeling via open-vocabulary detection.
\section*{Ethics Statement}
This work primarily focuses on autonomous driving data representation and reconstruction. Accordingly, we use open datasets captured in public spaces that strive to preserve personal privacy by leveraging state-of-the-art object detection techniques to blur people's faces and vehicle license plates. However, these are instance-level characteristics. What requires more effort to manage (and could potentially lead to greater harm) is maintaining a diversity of neighborhoods, and not only in terms of geography, but also population distribution, architectural diversity, and data collection times (ideally repeated traversals uniformly distributed throughout the day and night, for example). We created the NOTR dataset with diversity in mind, hand-picking scenarios from the Waymo Open Dataset \citep{sun2020scalability} to ensure a diversity of neighborhoods and scenario types (e.g., static, dynamic). However, as in the parent Waymo Open Dataset, the NOTR dataset contains primarily urban geographies, collected from only a handful of cities in the USA.
 
\section*{Reproducibility Statement}
We present our method in \S\ref{sec:method}, experiments and results in \S\ref{sec:experiment}, implementation details and ablation studies in \cref{app:implementation}. We benchmark previous approaches and our proposed method using publicly-available data and include details of the derived dataset in \cref{app:dataset}. Additional visualizations, code, models, and data are anonymously available either in the appendix or at {\small\url{https://emernerf.github.io}}.

\bibliography{iclr2024_conference}
\bibliographystyle{iclr2024_conference}

\clearpage
\appendix

\section{Implementation Details}
\label{app:implementation}

\renewcommand\thefigure{\thesection.\arabic{figure}}    
\setcounter{figure}{0}  
\renewcommand\thetable{\thesection.\arabic{table}}    
\setcounter{table}{0}  
\renewcommand\theequation{\thesection\arabic{equation}}    
\setcounter{equation}{0}  

In this section, we discuss the implementation details of \method. Our code is publicly available, and the pre-trained models will be released upon request. See \url{emernerf.github.io} for more details.

\subsection{\method implementation details}\label{app:specific_impl_details}

\subsubsection{Data Processing}

\paragraph{Data source.}  Our sequences are sourced from the \texttt{waymo\_open\_dataset\_scene\_flow}\footnote{\url{console.cloud.google.com/storage/browser/waymo_open_dataset_scene_flow}} version, which augments raw sensor data with point cloud flow annotations.  For camera images, we employ three frontal cameras: \texttt{FRONT\_LEFT}, \texttt{FRONT}, and \texttt{FRONT\_RIGHT}, resizing them to a resolution of $640\times960$ for both training and evaluation. Regarding LiDAR point clouds, we exclusively use the first return data (ignoring the second return data). We sidestep the rolling shutter issue in LiDAR sensors for simplicity and leave it for future exploration. Dynamic object masks are derived from 2D ground truth camera bounding boxes, with velocities determined from the given metadata. Only objects exceeding a velocity of $1$ m/s are classified as dynamic, filtering out potential sensor and annotation noise. For sky masks, we utilize the Mask2Former-architectured ViT-Adapter-L model pre-trained on ADE20k. Note that, the dynamic object masks and point cloud flows are used for \emph{evaluation only}.

\paragraph{Foundation model feature extraction.} We employ the officially released checkpoints of DINOv2 \cite{oquab2023dinov2} and DINOv1 \citep{dinov1}, in conjunction with the feature extractor implementation from \cite{amir2021deep}. For DINOv1, we utilize the ViT-B/16, resizing images to $640\times960$ and modifying the model's stride to 8 to further increase the resolution of extracted feature maps. For DINOv2, we use the ViT-B/14 variant, adjusting image dimensions to $644\times966$ and using a stride of 7. Given the vast size of the resultant feature maps, we employ PCA decomposition to reduce the feature dimension from 768 to 64 and normalize these features to the [0, 1] range.

\subsubsection{EmerNeRF}

 \paragraph{Representations.} We build all our scene representations based on iNGP \citep{muller2022instant} from \texttt{tiny-cuda-nn} \citep{tiny-cuda-nn}, and use \texttt{nerfacc} toolkit \citep{li2023nerfacc} for acceleration. Following \cite{barron2023zip}, our static hash encoder adopts a resolution spanning $2^4$ to $2^{13}$ over 10 levels, with a fixed feature length of 4 for all hash entries. Features at each level are capped at $2^{20}$ in size. With these settings, our model comprises approximately 30M parameters --- a saving of 18M compared to the StreetSurf's SDF representation \citep{guo2023streetsurf}.  For our dynamic hash encoder, we maintain a similar architecture, but with a maximum hash feature map size of $2^{18}$. Our flow encoder is identical to the dynamic encoder. To address camera exposure variations in the wild, 16-dimensional appearance embeddings are applied per image for scene reconstruction and per camera for novel view synthesis. While our current results are promising, we believe that larger hash encoders and MLPs could further enhance performance. See our code for more details.

 \paragraph{Positional embedding (PE) patterns.} We use a learnable feature map, denoted as $\mathcal{U}$, with dimensions $80\times 120 \times 32$ $(H\times W\times C)$  to accommodate the positional embedding patterns, as discussed in the main text. To decode the PE pattern for an individual pixel located at $(u, v)$, we first sample a feature vector from $\mathcal{U}$ using \texttt{F.grid\_sample}. Subsequently, a linear layer decodes this feature vector to produce the final PE features.

\paragraph{Scene range.} To define the axis-aligned bounding box (AABB) of the scene, we utilize LiDAR points. In practice, we uniformly subsample the LiDAR points by a factor of 4 and find the scene boundaries by computing the 2\% and 98\% percentiles of the world coordinates within the LiDAR point cloud. However, the LiDAR sensor typically covers only a 75-meter radius around the vehicle. Consequently, an unrestricted contraction mechanism is useful to ensure better performance. Following the scene contraction method detailed in \cite{reiser2023merf}, we use a piecewise-projective contraction function to project the points falling outside the determined AABB.

\paragraph{Multi-level sampling.} In line with findings in \cite{mildenhall2021nerf,barron2021mip}, we observe that leveraging extra proposal networks enhances both rendering quality and geometry estimation. Our framework integrates a two-step proposal sampling process, using two distinct iNGP-based proposal models. In the initial step, 128 samples are drawn using the first proposal model which consists of 8 levels, each having a 1-dim feature. The resolution for this model ranges from $2^{4}$ to $2^{9}$, and each level has a maximum hash map capacity of $2^{20}$. For the subsequent sampling phase, 64 samples are taken using the second proposal model. This model boasts a maximum resolution of $2^{11}$, but retains the other parameters of the first model. To counteract the ``z-aliasing'' issue---particularly prominent in driving sequences with thin structures like traffic signs and light poles, we further incorporate the anti-aliasing proposal loss introduced by \cite{barron2023zip} during proposal network optimization. A more thorough discussion on this is available in \cite{barron2023zip}. Lastly, we do not employ spatial-temporal proposal networks, i.e., we don't parameterize the proposal networks with a temporal dimension. Our current implementation already can capture temporal variations from the final scene fields, and we leave integrating a temporal dimension in proposal models for future exploration. For the final rendering, 64 points are sampled from the scene fields.

\paragraph{Pixel importance sampling.} Given the huge image sizes, we prioritize hard examples for efficient training. Every 2k steps, we render RGB images at a resolution reduced by a factor of 32 and compute the color errors against the ground truths. For each training batch, 25\% of the training rays are sampled proportionally based on these color discrepancies, while the remaining 75\% are uniformly sampled. This strategy is similar to \cite{wang2023tracking} and \cite{guo2023streetsurf}.

\subsubsection{Optimization}

All components in \method are trained jointly in an end-to-end manner.

\paragraph{Loss functions.} As we discussed in \S \ref{sec:optimization}, our total loss function is 
\begin{equation}
    \mathcal{L} = \underbrace{\mathcal{L}_{\text{rgb}} + \mathcal{L}_{\text{sky}} + \mathcal{L}_{\text{shadow}} + \mathcal{L}_{\sigma_d\text{(pixel)}} + \mathcal{L}_{\text{cycle}} + \mathcal{L}_{\text{feat}}}_{\text{for pixel rays}} + \underbrace{\mathcal{L}_{\text{depth}} + \mathcal{L}_{\sigma_d(\text{LiDAR})}}_{\text{for LiDAR rays}} %
\end{equation}
With $r$ representing a ray and $N_{r}$ its total number, the individual loss components are defined as:
\begin{enumerate}
    \item \textbf{RGB loss} ($\mathcal{L}_{\text{rgb}}$): Measures the difference between the predicted color ($\hat{C}(r)$) and the ground truth color ($C(r)$) for each ray.
    \begin{equation}
        \mathcal{L}_{\text{rgb}} = \frac{1}{N_{r}} \sum_{r} || \hat{C}(r) - C(r) ||^2_2
    \end{equation}
    \item  \textbf{Sky loss} ($\mathcal{L}_{\text{sky}}$): Measures the discrepancy between the predicted opacity of rendered rays and the actual sky masks. Specifically, sky regions should exhibit transparency. The binary cross entropy (BCE) loss is employed to evaluate this difference. In the equation, $\hat{O}(r)$ is the accumulated opacity of ray $r$ as in \Cref{eq:render}. $M(r)$  is the ground truth mask with 1 for the sky region and 0 otherwise.
    \begin{equation}
        \mathcal{L}_{\text{sky}} = 0.001 \cdot \frac{1}{N_{r}}\sum_{r}\mathrm{BCE}\left(\hat{O}(r), 1 - M(r) \right)
    \end{equation}
    \item \textbf{Shadow loss} ($\mathcal{L}_{\text{shadow}}$): Penalizes the accumulated squared shadow ratio, following \cite{wu2022d}.
    \begin{equation}
        \mathcal{L}_{\text{shadow}} = 0.01\cdot \frac{1}{N_{r}} \sum_{r}\left(\sum_{i=1}^{K} T_i \alpha_i \rho_i^2\right)
    \end{equation}
    \item \textbf{Dynamic regularization} ($\mathcal{L}_{\sigma_d(\text{pixel})}$ and $\mathcal{L}_{\sigma_d (\text{LiDAR})}$): Penalizes the mean dynamic density of all points across all rays. This encourages the dynamic branch to generate density only when necessary.
    \begin{equation}
        \mathcal{L_{\sigma_d}} = 0.01 \cdot \frac{1}{N_{r}} \sum_{r} \frac{1}{K} \sum_{i=1}^{K}\sigma_{d}(r, i)
    \end{equation}
    \item \textbf{Cycle consistency regularization} ($\mathcal{L}_{cycle}$): Self-regularizes the scene flow prediction. This loss encourages the congruence between the forward scene flow at time $t$ and its corresponding backward scene flow at time $t+1$.
    \begin{equation}
        \mathcal{L}_{\text{cycle}} = \frac{0.01}{2} \mathbb{E}\left[ \left[\mathrm{sg}(\mathbf{v}_f(\mathbf{x}, t)) + \mathbf{v}_b^{\prime}\left(\mathbf{x} + \mathbf{v}_f(\mathbf{x}, t), t + 1\right) \right]^2  \left[\mathrm{sg}(\mathbf{v}_b(\mathbf{x}, t)) + \mathbf{v}_f^{\prime}(\mathbf{x} + \mathbf{v}_b(\mathbf{x}, t), t-1) \right]^2 \right]
    \end{equation}
    where $\mathbf{v}_f(\mathbf{x}, t)$ denotes forward scene flow at time $t$, $\mathbf{v}_b^{\prime}(\mathbf{x} + \mathbf{v}_f(\mathbf{x}, t), t + 1)$ is predicted backward scene flow at the forward-warped position at time $t+1$, $\mathrm{sg}$ means stop-gradient operation, and $\mathbb{E}$ represents the expectation, i.e., averaging over all sample points.
    
    \item \textbf{Feature loss} ($\mathcal{L}_{\text{feat}}$): Measures the difference between the predicted semantic feature ($\hat{F}(r)$) and the ground truth semantic feature ($F(r)$) for each ray.
    \begin{equation}
    \mathcal{L}_{\text{feat}} = 0.5 \cdot \frac{1}{N_{r}}|| \hat{F}(r) - F(r) ||^2_2
    \end{equation}

    \item \textbf{Depth Loss} ($\mathcal{L}_{\text{depth}}$): Combines the expected depth loss and the line-of-sight loss, as described in \cite{rematas2022urban}. The expected depth loss ensures the depth predicted through the volumetric rendering process aligns with the LiDAR measurement's depth. The line-of-sight loss includes two components: a free-space regularization term that ensures zero density for points before the LiDAR termination points and a near-surface loss promoting density concentration around the termination surface. With a slight notation abuse, we have:
    \begin{align}
        \mathcal{L}_{\text{exp\_depth}} &= \mathbb{E}_{r}\left[|| \hat{Z}(r) - Z(r) ||_2^{2}\right] \\
        \mathcal{L}_{\text{line-of-sight}} &= \mathbb{E}_{r}\left[\int_{t_n}^{Z(r)-\epsilon} w(t)^2 dt\right] + \mathbb{E}_{r} \left[\int_{Z(r)-\epsilon}^{Z(r)+\epsilon}\left(w(t) - \mathcal{K}_{\epsilon}\left(t-Z(r)\right)\right)^2 \right] \\
        \mathcal{L}_{\text{depth}} &= \mathcal{L}_{\text{exp\_depth}} + 0.1 \cdot \mathcal{L}_{\text{line-of-sight}}
    \end{align}
    where $\hat{Z}(r)$ represents rendered depth values and $Z(r)$ stands for the ground truth LiDAR range values. Here, the variable $t$ indicates an offset from the origin towards the ray's direction, differentiating it from the temporal variable $t$ discussed earlier. $w(t)$ specifies the blending weights of a point along the ray. $\mathcal{K}\epsilon(x)=\mathcal{N}(0, (\epsilon/3)^2)$ represents a kernel integrating to one, where $\mathcal{N}$ is a truncated Gaussian. The parameter $\epsilon$ determines the strictness of the line-of-sight loss. Following the suggestions in \cite{rematas2022urban}, we linearly decay $\epsilon$ from 6.0 to 2.5 during the whole training process.
        
\end{enumerate}

\paragraph{Training.} We train our models for 25k iterations using a batch size of 8196. In static scenarios, we deactivate the dynamic and flow branches. Training durations on a single A100 GPU are as follows: for static scenes, feature-free training requires 33 minutes, while the feature-embedded approach takes 40 minutes. Dynamic scene training, which incorporates the flow field and feature aggregation, extends the durations to 2 hours for feature-free and 2.25 hours for feature-embedded representations. To mitigate excessive regularization when the geometry prediction is not reliable, we enable line-of-sight loss after the initial 2k iterations and subsequently halve its coefficient every 5k iterations.

\subsection{Basline implementations}
\label{app:baselines}

For HyperNeRF \citep{park2021hypernerf} and D$^2$NeRF \citep{wu2022d}, we modify their official JAX implementations to fit our NOTR dataset. Both models are trained for 100k iterations with a batch size of 4096. Training and evaluation for each model take approximately 4 hours on 4 A100 GPUs per scene. To ensure comparability, we augment both models with a sky head and provide them with the same depth and sky supervision as in our model. However, since neither HyperNeRF nor D$^2$NeRF inherently supports separate sampling of pixel rays and LiDAR rays, we project LiDAR point clouds onto the image plane and apply an L2 loss between predicted depth and rendered depth. We compute a scale factor from the AABBs derived from LiDAR data to ensure scenes are encapsulated within their predefined near-far range. For StreetSuRF \citep{guo2023streetsurf}, we adopt their official implementation but deactivate the monocular ``normal'' supervision for alignment with our setup. Additionally, to ensure both StreetSuRF and \method use the same data, we modify their code to accommodate our preprocessed LiDAR rays.

\section{NeRF On-The-Road (NOTR) Dataset}
\label{app:dataset}

\renewcommand\thefigure{\thesection.\arabic{figure}}    
\setcounter{figure}{0}  
\renewcommand\thetable{\thesection.\arabic{table}}    
\setcounter{table}{0} 

\begin{figure}
    \centering
    \includegraphics[width=1\linewidth]{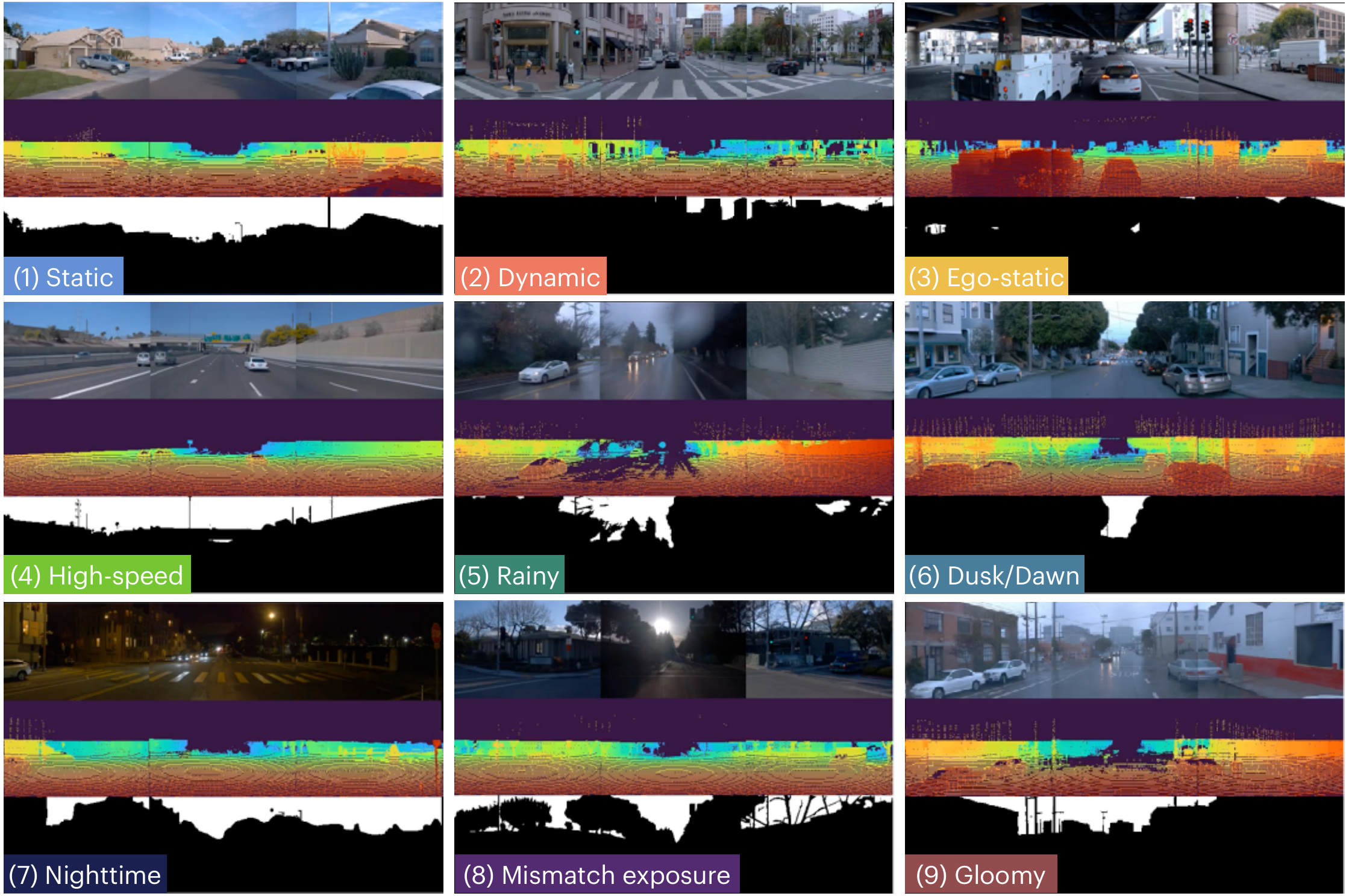}
    \caption{Samples from the NOTR Benchmark. This comprehensive benchmark contains (1) 32 static scenes, (2) 32 dynamic scenes, and 56 additional scenes across seven categories: (3) ego-static, (4) high-speed, (5) rainy, (6) dusk/dawn, (7) nighttime, (8) mismatched exposure, and (9) gloomy conditions. We include LiDAR visualization in each second row and sky masks in each third row.}
    \label{fig:dataset}
\end{figure}

As neural fields gain more attention in autonomous driving, there is an evident need for a comprehensive dataset that captures diverse on-road driving scenarios for NeRF evaluations. To this end, we introduce \textbf{N}eRF \textbf{O}n-\textbf{T}he-\textbf{R}oad (NOTR) dataset, a benchmark derived from the Waymo Open Dataset \citep{sun2020scalability}. NOTR features 120 unique driving sequences, split into 32 static scenes, 32 dynamic scenes, and 56 scenes across seven challenging conditions: ego-static, high-speed,  exposure mismatch, dusk/dawn, gloomy, rainy, and nighttime. Examples are shown in \Cref{fig:dataset}. 

Beyond images and point clouds, NOTR provides additional resources pivotal for driving perception tasks: bounding boxes for dynamic objects, ground-truth 3D scene flow, and 3D semantic occupancy. We hope this dataset can promote NeRF research in driving scenarios, extending the applications of NeRFs from mere view synthesis to motion understanding, e.g., 3D flows, and scene comprehension, e.g., semantics.

Regarding scene classifications, our static scenes adhere to the split presented in StreetSuRF \citep{guo2023streetsurf}, which contains clean scenes with no moving objects. The dynamic scenes, which are frequently observed in driving logs, are chosen based on lighting conditions to differentiate them from those in the ``diverse'' category. The Diverse-56 samples may also contain dynamic objects, but they are split primarily based on the ego vehicle's state (e.g., ego-static, high-speed, camera exposure mismatch), weather condition (e.g., rainy, gloomy), and lighting difference (e.g., nighttime, dusk/dawn). We provide the sequence IDs of these scenes in our codebase.

\section{Additional results}
\renewcommand\thefigure{\thesection.\arabic{figure}}    
\setcounter{figure}{0}  
\renewcommand\thetable{\thesection.\arabic{table}}    
\setcounter{table}{0}  
\subsection{Qualitative results}
\label{app:qualitative_comparison}

\begin{figure}
    \centering
    \includegraphics[width=0.8\linewidth]{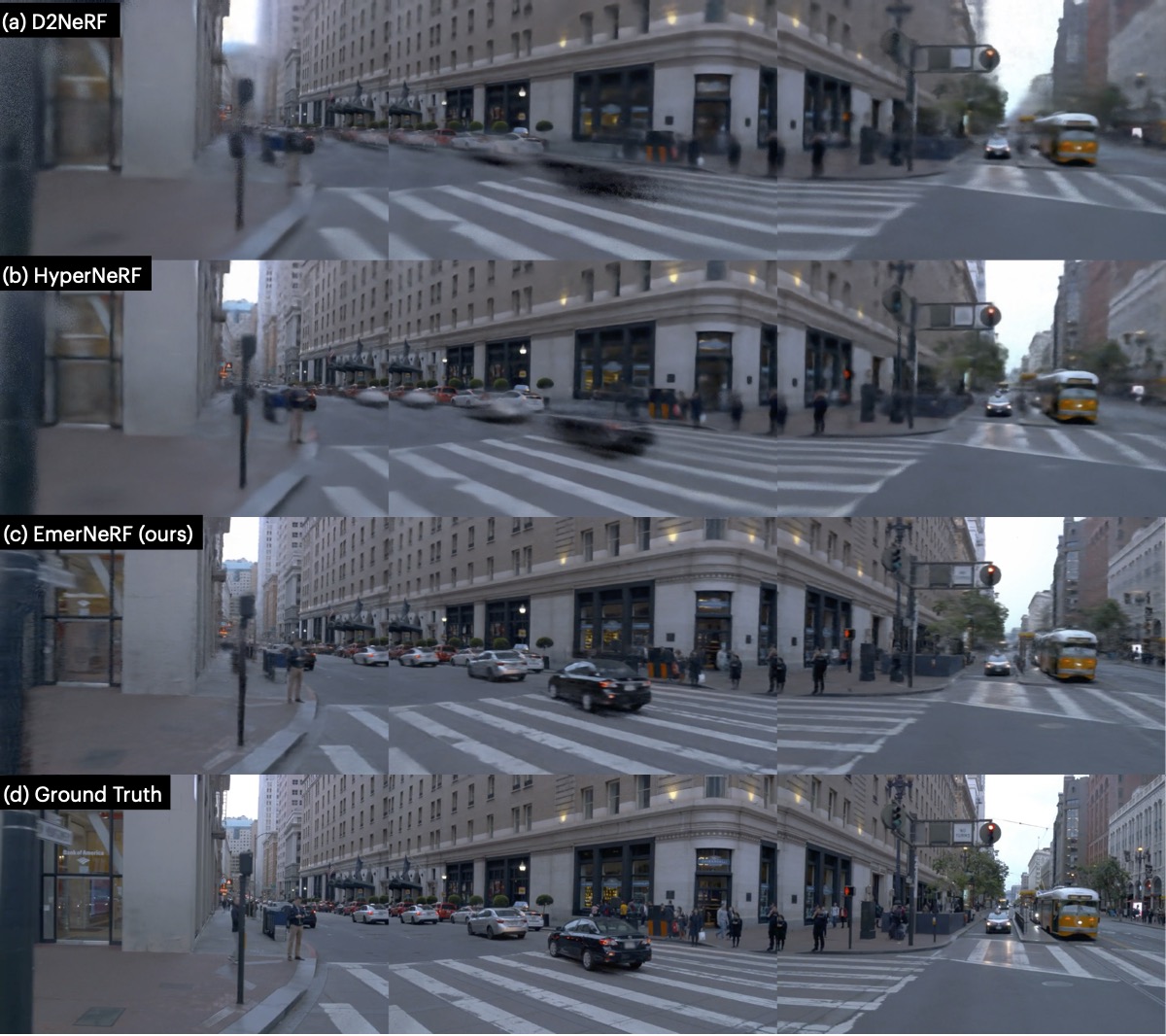}
    \caption{Qualitative scene reconstruction comparisons.}
    \label{fig:comparison1}
\end{figure}

\begin{figure}
    \centering
    \includegraphics[width=0.7\linewidth]{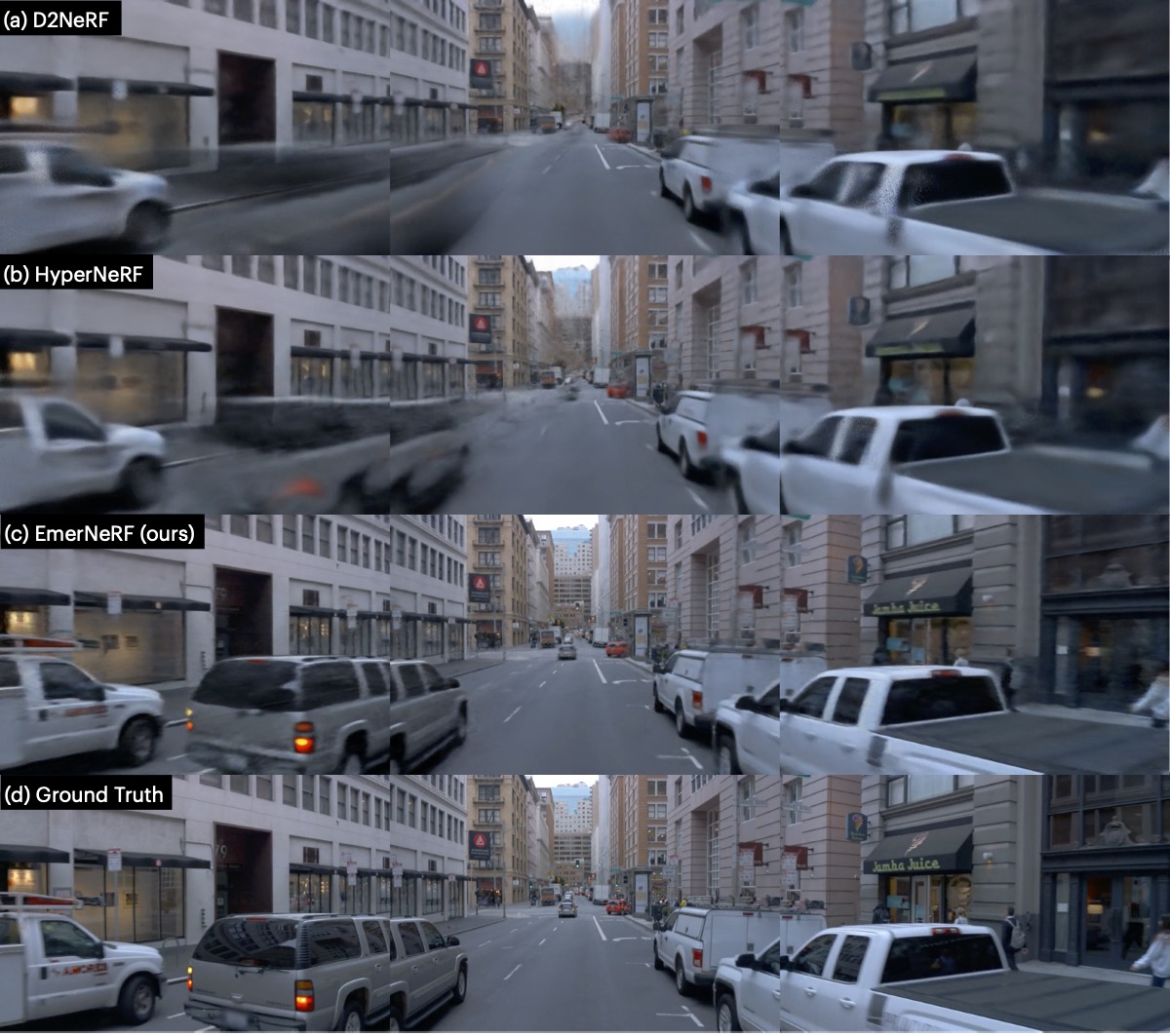}
    \caption{Qualitative scene reconstruction comparisons.}
    \label{fig:comparison2}
\end{figure}

\begin{figure}[h!]
    \centering
    \includegraphics[width=0.8\linewidth]{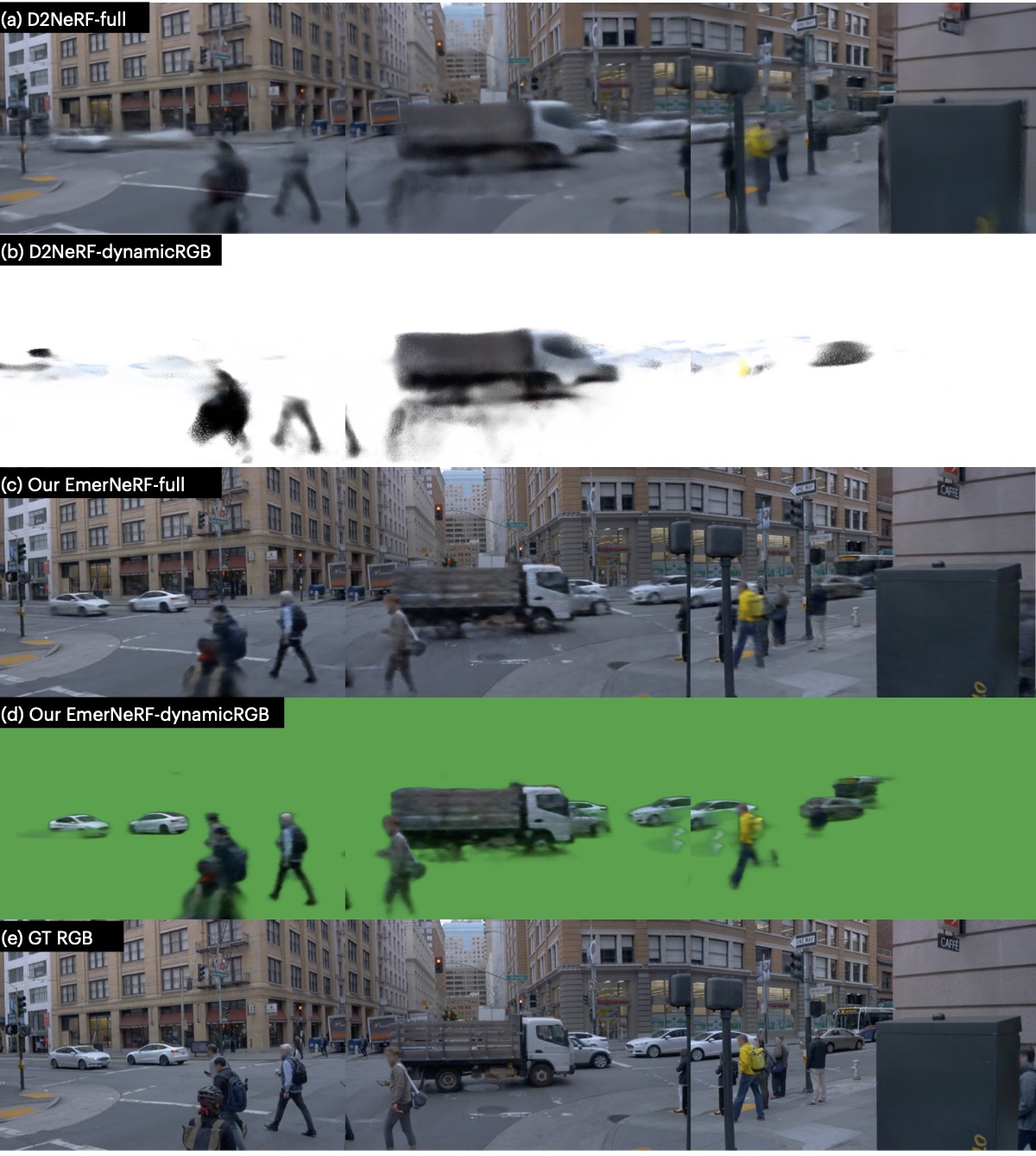}
    \caption{Scene decomposition comparisons. Note that we utilize a green background to blend dynamic objects, whereas D$2$NeRF's results are presented with a white background.}
    \label{fig:decomposition_comparison}
\end{figure}

\begin{figure}
    \centering
    \includegraphics[width=0.7\linewidth]{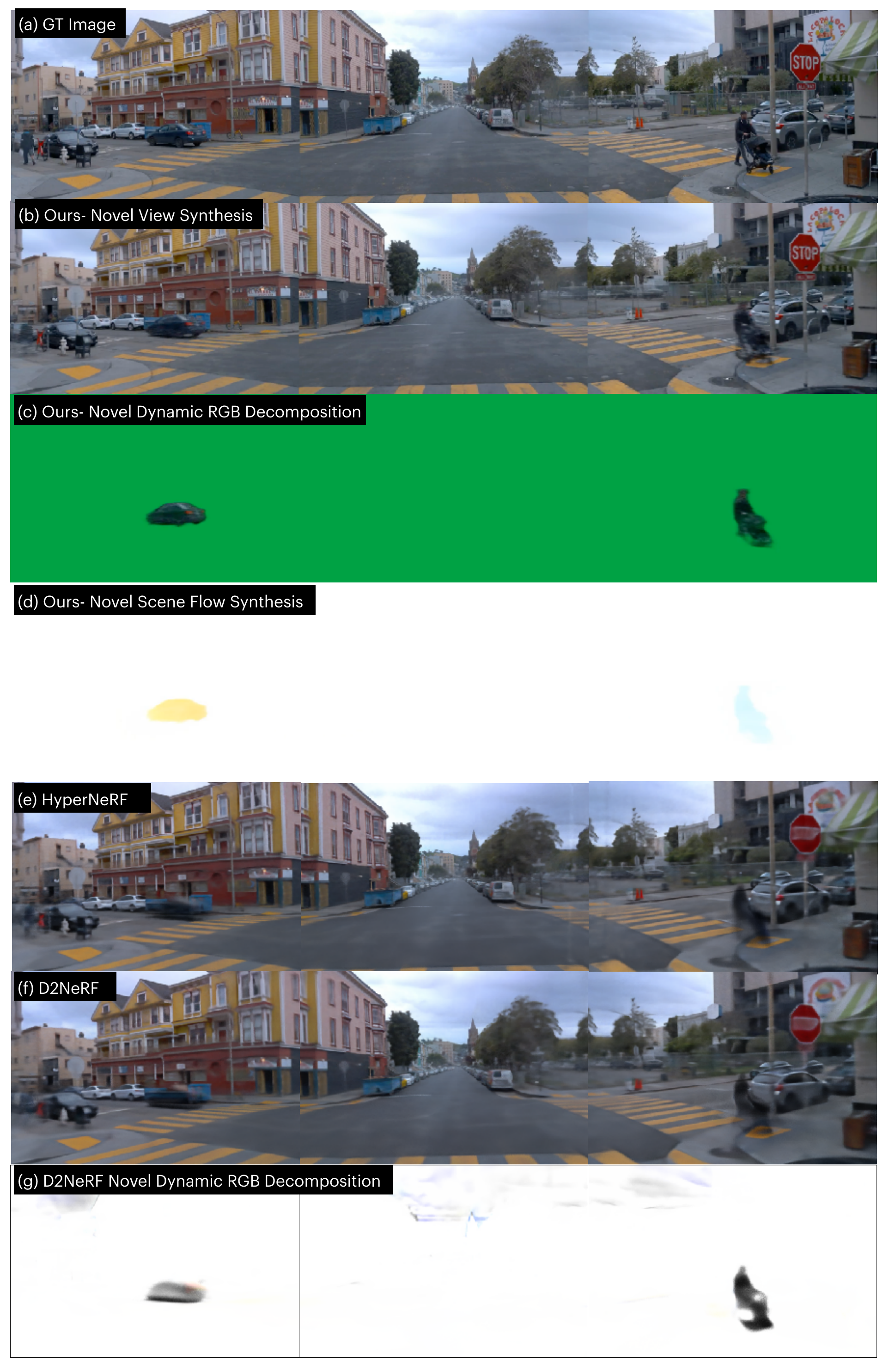}
    \caption{Qualitative novel temporal view comparison.}
    \label{fig:dynamic_nvs}
\end{figure}

\paragraph{Qualitative comparison.} \Cref{fig:comparison1,fig:comparison2} show qualitative comparisons between our \method and previous methods under the scene reconstruction setting, while \Cref{fig:decomposition_comparison} highlights the enhanced static-dynamic decomposition of our method compared to D$^2$NeRF \citep{wu2022d}. Moreover, \Cref{fig:dynamic_nvs} illustrates our method's superiority in novel view synthesis tasks against HyperNeRF \citep{park2021hypernerf} and D$^2$NeRF \citep{wu2022d}. Our method consistently delivers more realistic and detailed renders. Notably, HyperNeRF does not decompose static and dynamic components; it provides only composite renders, while our method not only renders high-fidelity temporal views but also precisely separates static and dynamic elements. Furthermore, our method introduces the novel capability of generating dynamic scene flows.

\subsection{Ablation Studies}
\label{app:ablation}

\Cref{tab:ablation} provides ablation studies to understand the impact of other components on scene reconstruction, novel view synthesis, and scene flow estimation. For these ablation experiments, all models are trained for 8k iterations, a shorter duration compared to the 25k iterations in the primary experiments. From our observations: (a) Using a full 4D iNGP without the static field results in the worst results, a consequence of the lack of multi-view supervision. (b-e) Introducing hybrid representations consistently improves the results. (c) Omitting the temporal aggregation step or (d) freezing temporal feature gradients (stop the gradients of $\mathbf{g}_d^{t-1}$ and $\mathbf{g}_d^{t+1}$ in \cref{fig:overview}) negates the emergence of flow estimation ability, as evidenced in the final column. Combining all these settings yields the best results.

\begin{table}[]
    \centering
    \caption{\textbf{Ablation study}.}
    \resizebox{\linewidth}{!}{%
    \begin{tabular}{lccccc}
    \toprule
    \multirow{3}{*}{Setting} & \multicolumn{2}{c}{Scene Reconstruction}      & \multicolumn{2}{c}{Novel View Synthesis}   & Scene Flow estimation  \\ 
    \cmidrule(lr){2-3} \cmidrule(lr){4-5} \cmidrule{6-6} 
     & Full Image     & Dynamic-Only & Full Image & Dynamic-Only & Flow \\
     & PSNR$\uparrow$  & PSNR$\uparrow$   & PSNR$\uparrow$ & PSNR$\uparrow$  & $\text{Acc}_{5} (\%)$ $\uparrow$    \\ 
    \midrule
    (a) 4D-Only iNGP & 26.55 & 22.30 & 26.02  & 21.03 & - \\
    (b) no flow & 26.92 & 23.82 & 26.33 & 23.81  & - \\
    (c) no temporal aggregation & 26.95 & 23.90 & 26.60  & 23.98 & 4.53\% \\
    (d) freeze temporally displaced features before aggregation & 26.93 & 24.02 & 26.78 & 23.81 & 3.87\% \\
    \midrule
    (e) ours default & \textbf{27.21} & \textbf{24.41}  & \textbf{26.93} & \textbf{24.07} & \textbf{89.74\%} \\

    \bottomrule
    \end{tabular}%
    }
    \label{tab:ablation}
\end{table}

\subsection{Limitations}
\label{app:limitations}

Consistent with other methods, \method does not optimize camera poses and is prone to rolling shutter effects of camera and LiDAR sensors. Future work to address this issue can investigate joint optimization of pixel-wise camera poses, and compensation for LiDAR rolling shutter alongside scene representations. Moreover, the balance between geometry and rendering quality remains a trade-off and needs further study. Lastly, \method occasionally struggles with estimating the motion of slow-moving objects when the ego-vehicle is moving fast---a challenge exacerbated by the limited observations. We leave these for future research.

\subsection{Visualizations}
\label{app:more_vis}

\begin{figure}
    \centering
    \includegraphics[width=1\linewidth]{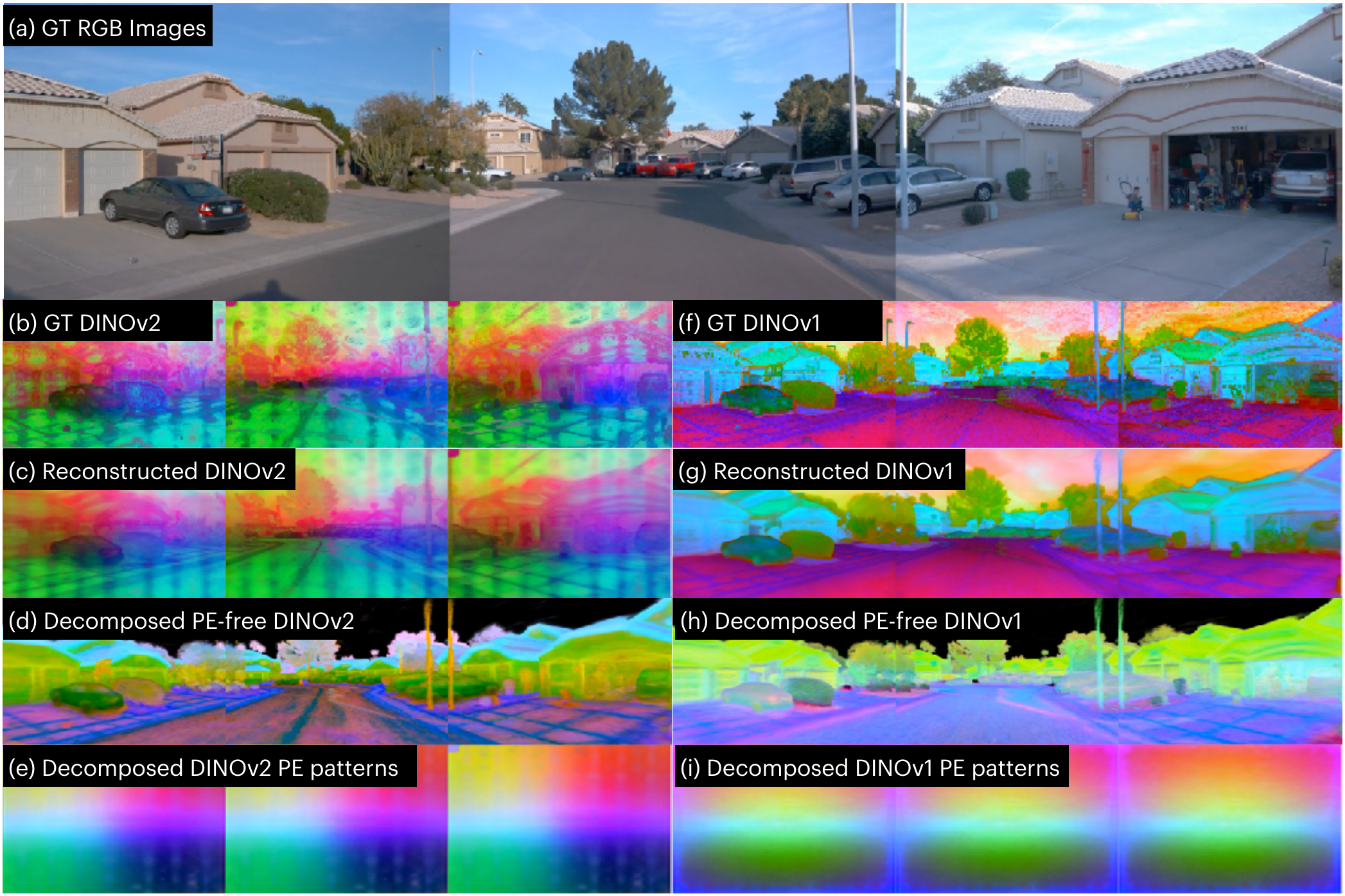}
    \caption{Different positional embedding patterns in DINOv1 \citep{dinov1} and DINOv2 models\citep{oquab2023dinov2}}
    \label{fig:pos_patterns}
\end{figure}

\begin{figure}
    \centering
    \includegraphics[width=1\linewidth]{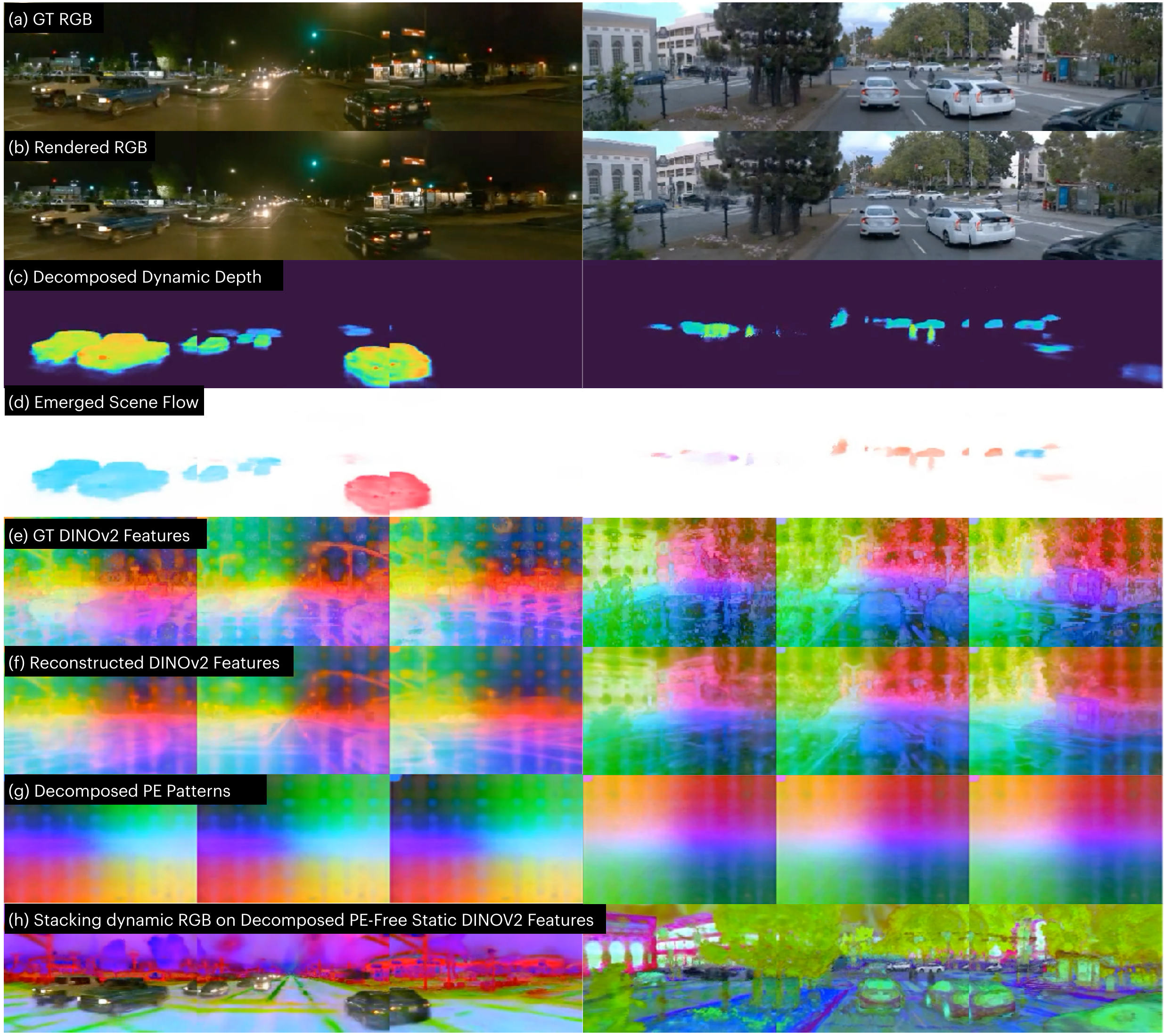}
    \caption{Scene reconstruction visualizations of \method. We show (a) GT RGB images, (b) reconstructed RGB images, (c) decomposed dynamic depth, (d) emerged scene flows, (e) GT DINOv2 features, (f) reconstructed DINOv2 features, and (g) decomposed PE patterns. (h) We also stack colors of dynamic objects onto decomposed PE-free static DINOv2 features.}
    \label{fig:example1}
\end{figure}
\begin{figure}
        \centering
        \includegraphics[width=1\linewidth]{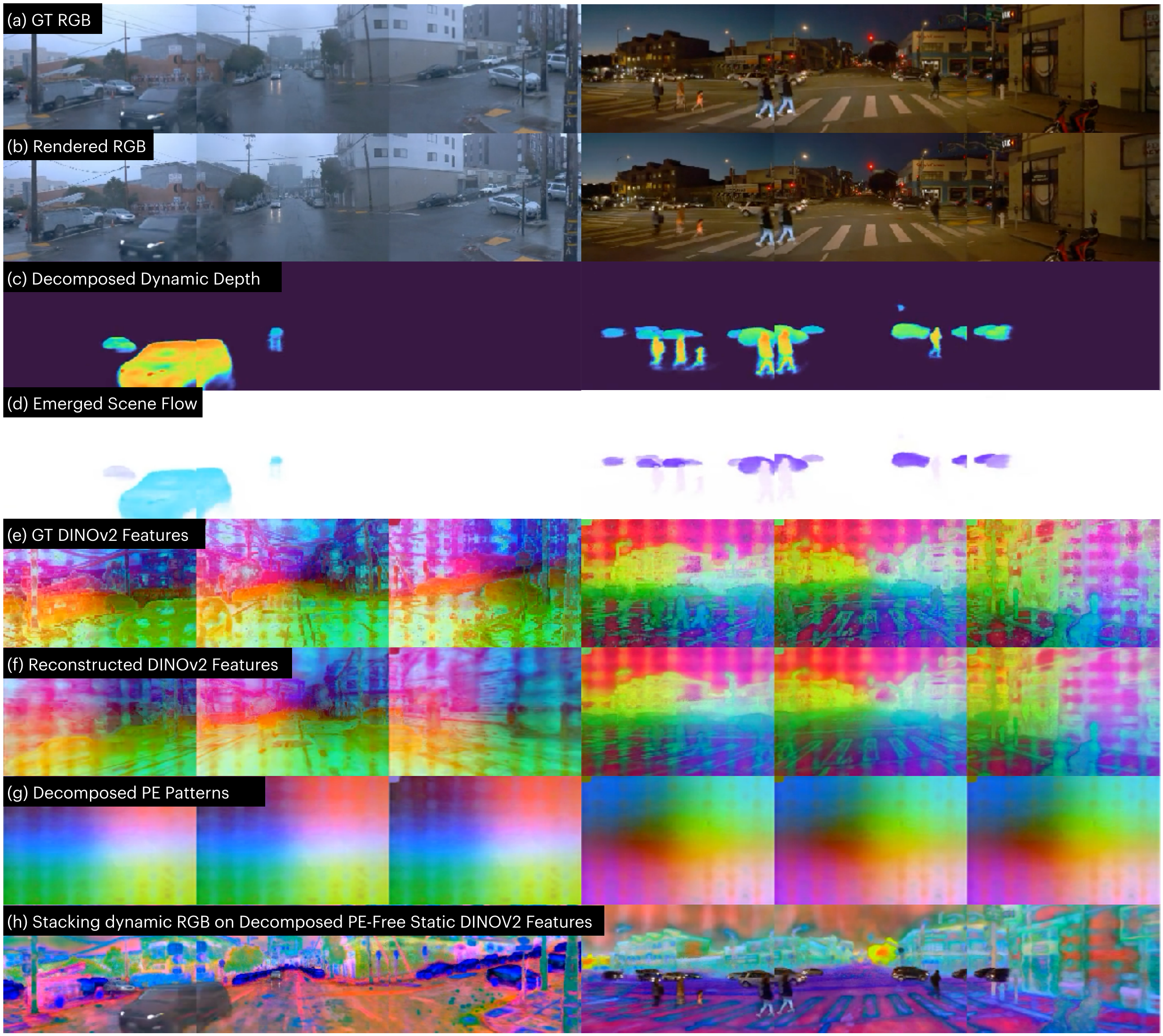}
        \caption{Scene reconstruction visualizations of \method under different lighting conditions. We show (a) GT RGB images, (b) reconstructed RGB images, (c) decomposed dynamic depth, (d) emerged scene flows, (e) GT DINOv2 features, (f) reconstructed DINOv2 features, and (g) decomposed PE patterns. (h) We also stack colors of dynamic objects onto decomposed PE-free static DINOv2 features. \method works well under dark environments (left) and discerns challenging scene flows in complex environments (right). Colors indicate scene flows' norms and directions.}
        \label{fig:example2}
    \end{figure}
\begin{figure}
    \centering
    \includegraphics[width=1\linewidth]{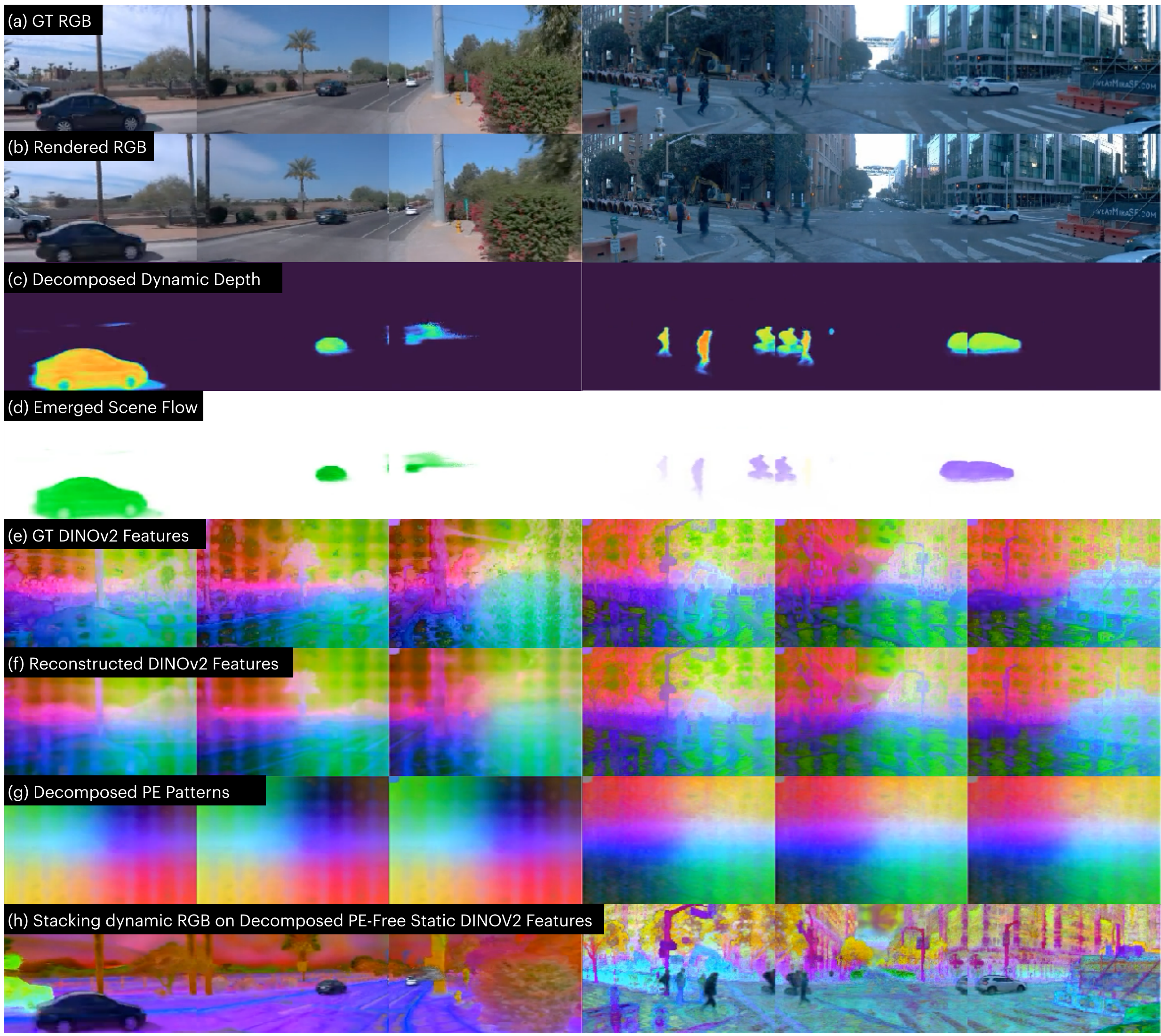}
    \caption{Scene reconstruction visualizations of \method under differet lighting and weather conditions. We show (a) GT RGB images, (b) reconstructed RGB images, (c) decomposed dynamic depth, (d) emerged scene flows, (e) GT DINOv2 features, (f) reconstructed DINOv2 features, and (g) decomposed PE patterns. (h) We also stack colors of dynamic objects colors onto decomposed PE-free static DINOv2 features. \method works well under gloomy environments (left) and discerns fine-grained speed information (right).}
    \label{fig:example3}
\end{figure}

\end{document}